\documentclass{article} 
\usepackage{iclr2025_conference,times}

\usepackage{amsmath,amsfonts,bm}

\def\eqref#1{equation~\ref{#1}}

\def\1{\bm{1}}

\DeclareMathAlphabet{\mathsfit}{\encodingdefault}{\sfdefault}{m}{sl}
\SetMathAlphabet{\mathsfit}{bold}{\encodingdefault}{\sfdefault}{bx}{n}

\usepackage{hyperref}
\hypersetup{
	colorlinks=true,
	citecolor=teal,
}
\usepackage{url}
\usepackage{booktabs}
\usepackage{microtype}      
\usepackage{multicol}
\usepackage{amsmath}
\usepackage{enumitem}
\usepackage{amssymb}
\numberwithin{equation}{section} 
\usepackage{wrapfig,lipsum,booktabs}
\usepackage{xcolor}
\usepackage{multirow}
\usepackage{float}
\usepackage{graphicx}
\usepackage{subcaption}
\usepackage{algpseudocode}  
\usepackage{algorithmicx,algorithm}
\usepackage{caption}
\usepackage{diagbox} 
\usepackage{colortbl}
\usepackage{xcolor}
\definecolor{rowcolor}{rgb}{0.9, 0.9, 0.9}
\usepackage{soul}

\usepackage{hyperref}
\usepackage{url}

\usepackage{algorithmicx,algorithm}
\definecolor{commentcolor}{RGB}{110,154,155}   
\definecolor{functioncolor}{RGB}{200,2,127}   

\usepackage{array} 
\usepackage{adjustbox}
\usepackage{endnotes}

\newcommand{\ie}[0]{\emph{i.e., }}

\newcommand{\eg}[0]{\emph{e.g., }}

\newcommand{\etc}[0]{\emph{etc.}}

\usepackage{graphicx}
\DeclareGraphicsExtensions{.pdf,.png,.jpg}

\pdfoutput=1
\usepackage{epstopdf}

\newcommand{\shortname}{\textit{TemporalBench}}

\title{\shortname{}: Benchmarking Fine-grained \\ Temporal Understanding for  Multimodal Video  Models}

 \iclrfinalcopy

\author{
   \hspace{-0.18cm}  \vspace{0.08cm}Mu Cai$^{1,}$\thanks{Work done during the internship at Microsoft Research,  $^{\dagger}$ Equal Advisory Contribution.}\hspace{0.15cm},\, Reuben Tan$^{2}$,\, Jianrui Zhang$^{1}$,\, Bocheng Zou$^{1}$,\, Kai Zhang$^{3}$,\, Feng Yao$^{4}$, \\
    \vspace{0.08cm}\textbf{Fangrui Zhu$^{5}$,\, Jing Gu$^{6}$,\, Yiwu Zhong$^{7}$,\, Yuzhang Shang$^{8}$,\, Yao Dou$^{9}$,\, Jaden Park$^{1}$,} \\
    \vspace{0.08cm}\textbf{Jianfeng Gao$^{2, \dagger}$,\, Yong Jae Lee$^{1,\dagger}$,\, Jianwei Yang$^{2,\dagger}$} \vspace{0.2cm} \\
             \footnotesize  $^{1}$University of Wisconsin-Madison    \hspace{.5in}   $^{2}$Microsoft Research, Redmond \vspace{0.1cm} \\
               \footnotesize $^{3}$ Ohio State University   \hspace{.1in}     $^{4}$  University of California, San Diego  \hspace{.1in}   $^{5}$ Northeastern University  \vspace{0.1cm}  \\
                 \footnotesize  $^{6}$  University of California, Santa Cruz  \hspace{.1in}   $^{7}$  Chinese University of Hong Kong   \hspace{.1in} \vspace{0.1cm}\\ 
                  \footnotesize  $^{8}$  Illinois Institute of Technology  \hspace{.1in}   $^{9}$  Georgia Institute of Technology  \vspace{0.2cm} \\
    \url{https://TemporalBench.github.io/}
}

\begin{document}

\maketitle

\begin{abstract}

Understanding fine-grained temporal dynamics is crucial for multimodal video comprehension and generation. Due to the lack of fine-grained temporal annotations, existing video benchmarks mostly resemble static image benchmarks and are incompetent at evaluating models for temporal understanding. In this paper, we introduce \shortname{}, a new benchmark dedicated to evaluating \textbf{fine-grained temporal understanding} in videos. \shortname{} consists of $\sim$10K video question-answer pairs, derived from $\sim$2K high-quality human annotations detailing the temporal dynamics in video clips. As a result, our benchmark provides a unique testbed for evaluating various temporal understanding and reasoning abilities such as \textit{action frequency, motion magnitude, event order,} \etc ~  Moreover, it enables evaluations on various tasks like both video question answering and captioning,  both short and long video understanding,  as well as different models such as multimodal video embedding models and text generation models. Results show that state-of-the-art models like GPT-4o achieve only $38.5\%$ question answering accuracy on \shortname{}, demonstrating a significant gap ($\sim30\%$) between humans and AI in temporal understanding.  Furthermore, we notice a critical pitfall for multi-choice QA where LLMs can detect the subtle changes in negative captions and find a ``centralized” description as a cue for its prediction, where we propose Multiple Binary Accuracy (MBA) to correct such bias. We hope that \shortname{} can foster research on improving models' temporal reasoning capabilities. Both dataset and evaluation code will be made available. 

%

\end{abstract}

\vspace{-10pt}
\section{Introduction}
\vspace{-5pt}

The ability to understand and reason about events in videos is a crucial aspect of artificial intelligence, with applications ranging from activity recognition and long-term action anticipation to perception for autonomous driving and robotics. Recently, there has been an emergence of highly capable multimodal generative models, including proprietary ones such as GPT-4o~\citep{GPT4o} and Gemini~\citep{geminiteam2024gemini} as well as open-sources ones~\citep{liu2023llava,zhu2023minigpt,Qwen-VL}, that have demonstrated impressive results on existing video benchmarks~\citep{xu2016msr,chen2011msvd,yu2019activitynet,mangalam2023egoschema}. However, these benchmarks often do not truly evaluate the abilities of the aforementioned models to understand video content due to their generally \emph{coarse-grained} annotations.

The lack of fine-grained temporal details in the annotations often leads to existing video understanding benchmarks suffering from a strong language prior bias. This is similar to observations in visual question answering with images~\citep{antol2015vqav1}. For example, prior works~\citep{tan2024koala, li2023seed} show that language models such as Flan-T5~\citep{chung2022scalinginstructionfinetunedlanguagemodels} and Llama-2/3~\citep{touvron2023llama} perform comparably to video models on EgoSchema~\citep{mangalam2023egoschema} and Seed-Bench~\citep{li2023seed} without using any information from videos. Furthermore, the lack of fine-grained temporal details often results in the single frame bias of current video understanding benchmarks~\citep{lei2022revealing}. These benchmarks are often biased toward spatial reasoning, where static information from a single frame suffices to achieve high performance. They often fail to test a model’s ability to reason about temporal sequences, leading to inflated evaluations of AI models that are not genuinely capable of understanding temporal events. Specifically, vision-language models (VLMs)~\citep{liu2023improvedllava, liu2024llavanext} that are trained on image-level datasets, including FreeVA~\citep{wu2024freeva}, IG-VLM~\citep{kim2024image} and $M^3$~\citep{cai2024matryoshka}, often outperform their video counterparts on popular video question answering benchmarks such as MSRVTT~\citep{xu2016msr}, MSVD~\citep{xu2017video}, and TGIF~\citep{jang2017tgif}. 

\begin{figure}[t]
    \centering
    \includegraphics[width=\linewidth]{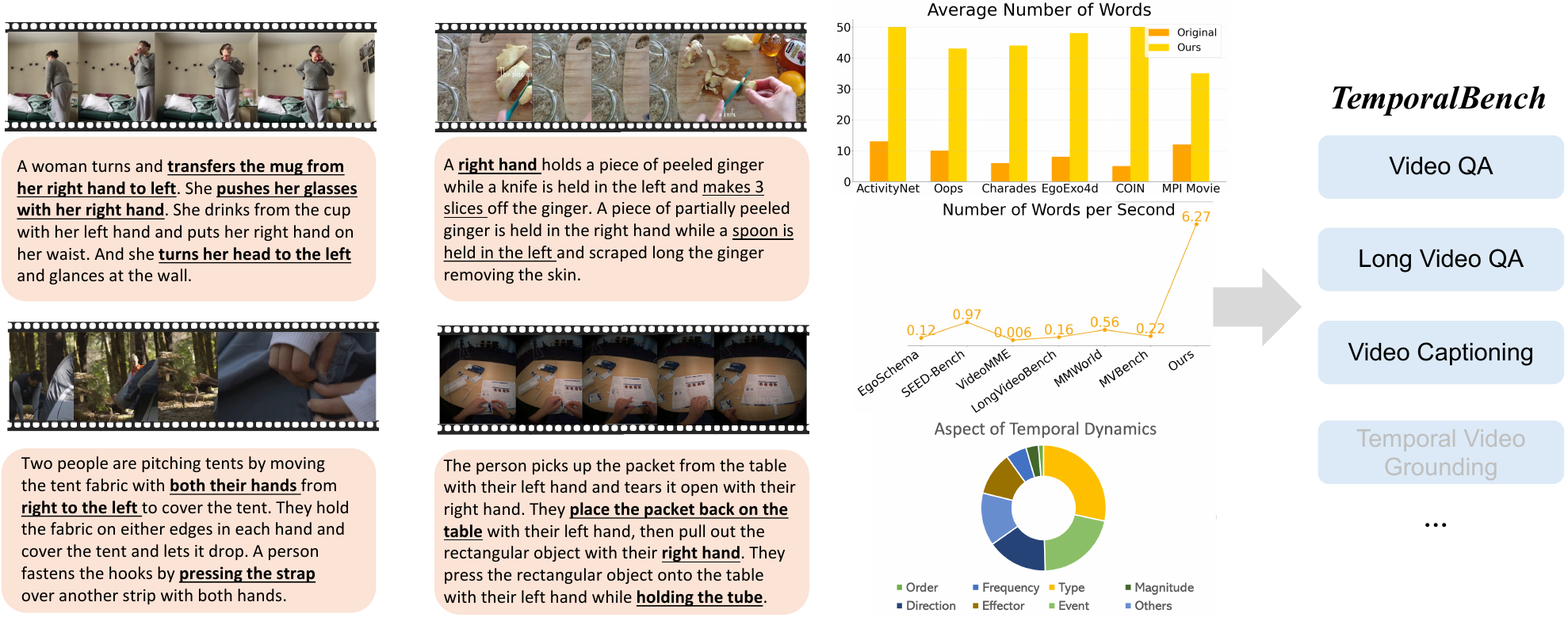}
    \caption{\textbf{The tasks of \shortname{}}. \shortname{} starts from fine-grained video descriptions and supports diverse video understanding tasks including video QA, video captioning, long video understanding, \textit{etc}. It differs from existing benchmarks by the average number of words per video (middle top), word density (center)  and the coverage of various temporal aspects (middle bottom).}
    \label{fig:teaser}
\end{figure}

To address this limitation, we propose ~\shortname{} (Figure~\ref{fig:teaser}), a new video understanding benchmark that evaluates multimodal video models on understanding fine-grained activities, and consists of \textbf{$\sim$10K} question and answer pairs curated from \textbf{$\sim$2K} high-quality human-annotated captions with rich activity details. Unlike static image-based tasks, video understanding requires models to reason effectively about both spatial and temporal information. The temporal dynamics inherent in videos introduce significant complexity, as actions and events often unfold over time and cannot be captured in a single frame. 

With this in mind, we designed our benchmark to focus on areas where current models often struggle, emphasizing annotations related to long-range dependencies, fine-grained visual observations, and event progression.

\begin{figure}[t]
    \centering
    \includegraphics[width=\linewidth]{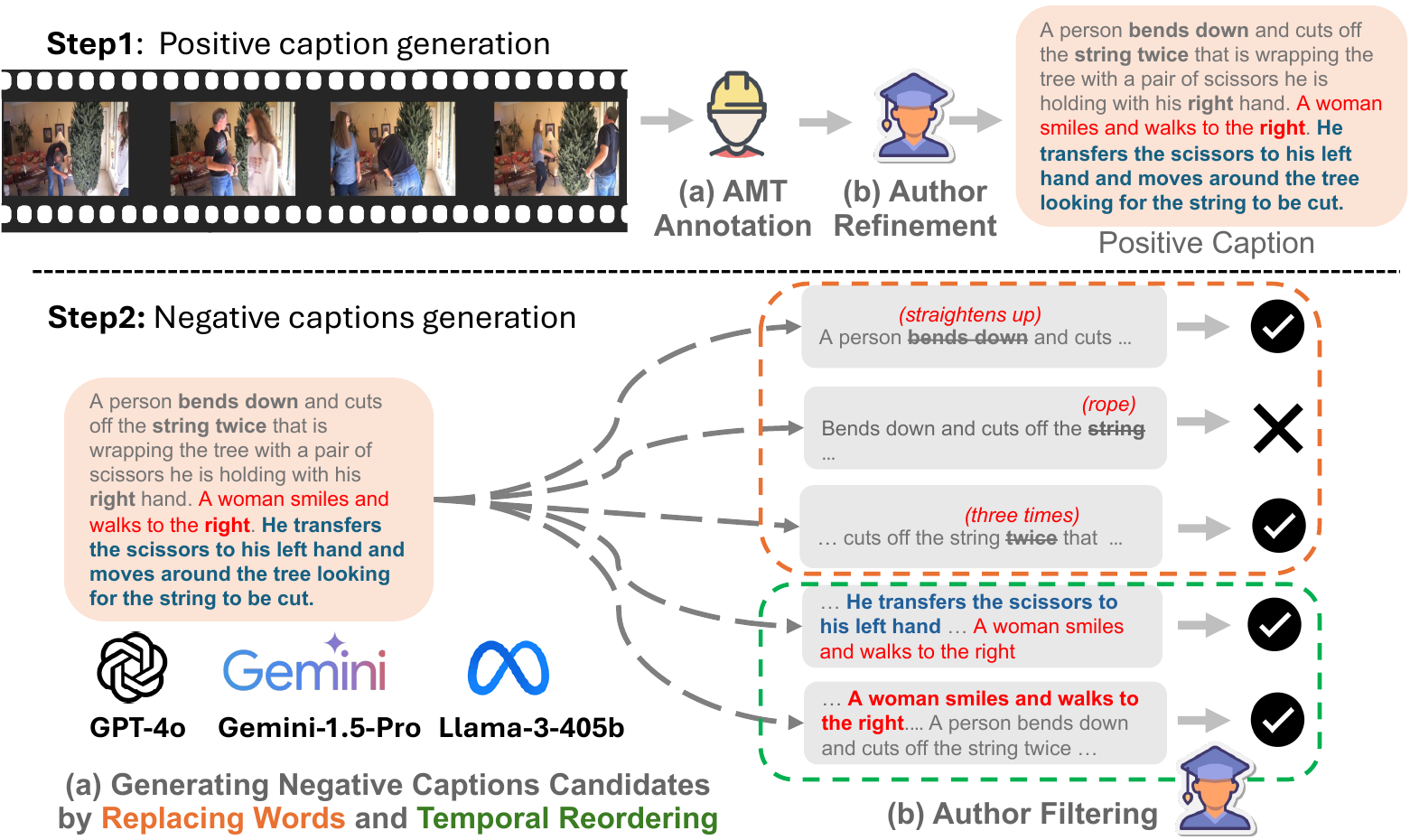}
    \caption{\textbf{Overview of the annotation pipeline for \shortname{}}. In step 1, we fist collect high-quality captions for the videos using qualified AMT annotators followed by refining them. In step 2, we leverage existing LLMs to generate negative captions by replacing select words and reordering the sequence of actions before filtering them ourselves.
    }
    \label{fig:benchmark_annotation}
\end{figure}

As shown in Figure~\ref{fig:benchmark_annotation}, we first collect video clips from existing video grounding benchmarks that span diverse domains, including procedural videos~\citep{tang2019coin}, human activities~\citep{krishna2017dense, gao2017tall}, ego-centric videos~\citep{grauman2024ego}, movie descriptions~\citep{rohrbach15cvpr}, professional gymnasium videos (FineGym from ~\cite{shao2020finegym}), and unexpected humor videos~\citep{oops}. The positive captions include \textit{rich} and \textit{fine-grained} details about actions and activities, which are annotated by highly qualified Amazon Mechanical Turk~(AMT) workers and authors of this paper. Then, we generate the negative captions with respect to the actions using powerful Large Language Models (LLMs) and filter them according to our defined rules. Our resulting \shortname{} contains $\sim$10K video descriptions and matching questions of high quality. Furthermore, the rich temporal context of annotations in our diverse corpus creates a solid foundation for the development of additional benchmarks in related tasks such as spatio-temporal localization and causal inference. We hope that our benchmark can pave the road for further development of multimodal video models capable of fine-grained video understanding and reasoning.

\begin{figure}[t]
    \centering
    \includegraphics[width=\linewidth]{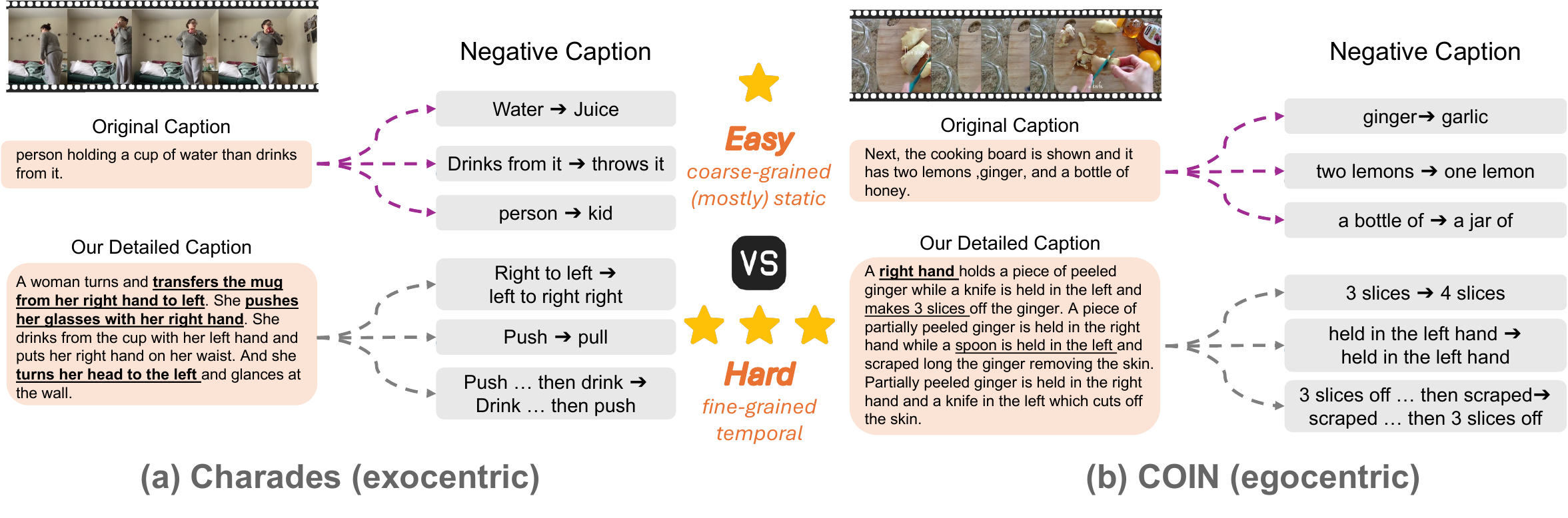}
    \caption{\textbf{Comparison of negative captions generated from the original captions and our detailed captions in \shortname{}}. With fine-grained details, the negatives are more difficult and temporal centric.
    }
    \label{fig:negative-caption-compare}
\end{figure}

In contrast to existing video benchmarks, \shortname{} has the following defining characteristics: 
\begin{itemize}[left=0pt]
    \item \textbf{Emphasis on fine-grained action understanding}. Due to the highly descriptive video captions, our negative captions highlight fine-grained temporal differences shown in Figure~\ref{fig:negative-caption-compare}, such as \textit{``sliced the ginger three times" }versus \textit{``sliced the ginger twice"},   and \textit{``put on the eyeglasses"} versus \textit{``push the eyeglasses"}.
    \item \textbf{Evaluations on both short ($<$20 seconds) and long ($<$20 minute) videos.} Since the videos clips are sampled from existing videos, our benchmark can also support evaluations on long video understanding by concatenating the descriptions of multiple and non-overlapping video clips from the same source video. 
    \item \textbf{Extends to video captioning, video grounding, and video generation.} Besides the task of video question answering, the nature of the positive captions in our benchmark allows it to seamlessly extend to evaluation of other tasks such as video temporal grounding and dense captioning.
    \item \textbf{Evaluations of both video embedding and question-answering models.} Given the annotated positive and negative captions in \shortname{}, it also supports the evaluation of discriminative and contrastive learning-based models such as XCLIP~\citep{xclip}, ImageBind~\citep{girdhar2023imagebind} as well as multimodal generative models such as GPT-4o and Gemini. 
\end{itemize} 

Furthermore, we notice a critical pitfall for multi-choice QA. If every negative answer choice is generated by changing a small part of the
correct answer, the LLM can detect those changes to find a “centralized” description and use that cue
for its prediction.  Therefore, we propose Multiple Binary Accuracy (MBA) to correct such bias.

Among other observations, our empirical evaluations show that state-of-the-art multimodal video models like GPT-4o only achieve an average accuracy of 38.5\% on our benchmark (short videos) using our proposed multiple binary QA accuracy metric, compared to {67.9\%} obtained by humans. Models show even worse results on long videos. 
This result highlights that the aforementioned models are able to understand static visual concepts but are still limited in reasoning about the fine-grained temporal relationships of objects and events in videos. More significantly, we highlight a critical issue with using LLMs to answer multi-choice QA.

\vspace{-5pt}
\section{Related Work}
\vspace{-5pt}

\textbf{Large Multimodal Models.}
Large Language Models (LLMs) like ChatGPT~\citep{chatgpt}, GPT-4~\citep{gpt4}, and Llama~\citep{touvron2023llama} have demonstrated impressive reasoning and generalization capabilities for text. The introduction of models that integrate visual data has brought about a significant shift in the landscape of LLMs, such as GPT-4V(ision)\citep{GPT4V_System_Card}. Building upon open-source LLMs \citep{touvron2023llama,vicuna2023}, a wide range of multimodal models has achieved remarkable progress, led by pioneering models such as LLaVA~\citep{liu2023llava, liu2023improvedllava} and MiniGPT-4~\citep{zhu2023minigpt}, which combine LLMs' capabilities with a CLIP~\citep{radford2021learning} based image encoder. Recently,  a growing number of LMMs have been developed to handle a wider range of tasks and modalities, such as region-level LMMs~\citep{cai2024vipllava, zhang2023gpt4roi, chen2023shikra, peng2023kosmos,zhang2023llavagrounding}, 3D LMMs~\citep{3dllm}, and video LMMs~\citep{lin2023video, zhang2023video, zhang2024llavanextvideo}.

\textbf{Multimodal Understanding Benchmarks.} The recent significant advancements have resulted in more versatile multimodal models, making it imperative to thoroughly and extensively evaluate their visual understanding and reasoning abilities. Conventional multimodal benchmarks like VQA~\citep{antol2015vqav1}, GQA~\citep{hudson2019gqa} and VizWiz~\citep{gurari2018vizwiz} have been revitalized and used for evaluating the general visual question answering performance for LMMs. Some other question answering benchmarks like TextVQA~\citep{singh2019textvqa}, DocVQA~\citep{mathew2021docvqa} and InfoVQA~\citep{mathew2021infographicvqa} have also been employed to validate the text-oriented understanding. Recent studies have introduced a variety of new benchmarks, such as SEED-Bench~\citep{li2023seed}, MMBench~\citep{liu2023mmbench} and MM-Vet~\citep{yu2023mmvet} for evaluating the models' integrated problem-solving capabilities, and MMMU~\citep{yue2023mmmu} and MathVista~\citep{lu2024mathvistaevaluatingmathematicalreasoning} for scientific and mathematical reasoning. In addition, the commonly known hallucination problem also appears in LMMs, and is also investigated in POPE~\citep{li2023pope}, MMHal-Bench~\citep{sun2023aligning} and Object HalBench~\citep{yu2024rlhf}, \textit{etc}. 

\textbf{Video Understanding Benchmarks.} Recently, an increasing amount of research is transitioning its focus from the image to the video domain. Videos differ from images in that they possess more complex content with temporal dynamics. This unique aspect calls for a different set of metrics and benchmarks. Many efforts have leveraged existing video question answering benchmarks~\citep{xu2017video,yu2019activitynetqadatasetunderstandingcomplex,xiao2021next} built on top of video-text datasets~\citep{chen2011msvd,xu2016msr,zhang2019dynamictemporalpyramidnetwork}. More recently, several LMM-oriented benchmarks have been proposed for different aspects such as long-form egocentric understanding with EgoSchema~\citep{mangalam2023egoschema}, and temporal understanding and ordering like Tempcompass~\citep{liu2024tempcompassvideollmsreally}. MV-Bench~\citep{li2024mvbenchcomprehensivemultimodalvideo} compiles existing video annotations from different disciplines into a new benchmark, while Video-MME~\citep{fu2024videommefirstevercomprehensiveevaluation} and MMWorld~\citep{he2024mmworldmultidisciplinemultifacetedworld} claim to support a comprehensive evaluation of video understanding and world modeling, respectively. Our \shortname{} serves the common goal of evaluating models for video understanding but differs in several aspects. On the one hand, we exhaustively curate videos from different domains and ask human annotators to annotate the visual contents with as much detail as possible. On the other hand, we particularly focus on temporal dynamics such as human actions and human-object interactions that exist exclusively in videos and which are crucial for video understanding, reasoning and forecasting. While the ShareGPT4Video dataset~\citep{chen2024sharegpt4video} also contains long captions, theirs differ from ours by being entirely generated by GPT-4o instead of annotated by humans.

\vspace{-5pt}
\section{\shortname{}}
\label{sec:approach}
\vspace{-5pt}

Compared to static images, videos inherently contain significantly more fine-grained temporal information, as they capture the unfolding of actions and events over time. Existing multimodal video understanding benchmarks~\citep{xu2016msr} mostly evaluate models' coarse-level understanding of videos. An example from the recent Seed-Bench dataset is the question, \textit{``What action is happening in the video?"} with the answer, \textit{``moving something up."} However, such types of coarse-level video questions have been demonstrated to be easily solved with just a single frame~\citep{wu2024freeva} or even by a text-only LLM~\citep{tan2024koala, mangalam2023egoschema}.

Such phenomena arises due to a fundamental limitation in the text descriptions in those benchmarks. As a result of their coarseness, the positive and negative options for video question-answering can usually be distinguished without understanding the temporal dynamics, such as the models only needing to choose between \textit{``The man is cooking"} and \textit{``The man is exercising"}.

To address this limitation, we carefully design a human annotation pipeline to curate highly detailed descriptions about the activities in the videos. Given the detailed video clip descriptions, such as \textit{A right hand holds a piece of peeled ginger while a knife is held in the left and makes 3 slices off the ginger.},  the negative captions can be curated to truly reflect whether a model understands the temporal dynamics, such as changing \textit{``three slices''} into \textit{``two slices''}. In a nutshell, such highly detailed temporal annotations can be used to carefully examine whether a multimodel video model truly  understands the temporal state transition in videos.

Our benchmark enriches several fundamental video understanding tasks due to its detailed captions: 

\begin{itemize}[left=0pt]
    \item \textbf{Fine-grained video question answering.} Given a detailed positive caption,  multimodal video models need to distinguish it from the associated negative where a slight modification is made to temporal descriptions, \eg \textit{``push the eyeglasses up''} versus \textit{``pull the eyeglasses down''}, or \textit{``cut 3 slices off''} versus \textit{``cut 2 slices off''}. 
    \item \textbf{Fine-grained video captioning.} Our detailed video captions can naturally enrich the video captioning task, different from current video captioning tasks such as MSRVTT~\citep{xu2016msr} which focus on coarse-level descriptions. 
    \item \textbf{Long video understanding with fine-grained activity inspection.} Since the video clips are extracted from a long source video, the respective video clip descriptions can be concatenated to form a longer video description which can be pivoted to the long video understanding task, where we find that all current multimodal video models suffer.
    \item \textbf{Dense video-text matching and retrieval.} Our detailed video captions can be naturally employed to evaluate video-language embedding models such as XCLIP~\citep{xclip}. Given a positive caption and several negative captions, we can evaluate whether CLIP~\citep{radford2021learning} based video embedding models can distinguish the subtle differences in captions. In addition, given a set of positive video-text pairs,  video retrieval performance can be evaluated, similar to image retrieval on COCO~\citep{lin2014microsoft} and Flickr30K~\citep{young2014image}. 
    \item \textbf{Video grounding from detailed text descriptions.} Since the video clips are cropped from the source video, with the documented starting and ending time, our benchmark can serve as a fine-grained moment localizing benchmark from text descriptions. This is different from existing video grounding datasets such as Charades-STA~\citep{gao2017tall}, COIN~\citep{tang2019coin}, Ego4D~\citep{grauman2024ego} where the text descriptions are usually very short, possibly resulting in low temporal localization performance due to the vague and coarse descriptions.
    \item \textbf{Text-to-Video (T2V) generation with detailed prompts.} Given our highly detailed description, a T2V generation model can be evaluated by verifying if the generated videos reflect the fine-grained action details.
\end{itemize}

Next, we detail the dataset curation and evaluation setup for \shortname.

\subsection{Video Collection}

We collect video clips from a wide range of sources across diverse domains, where the majority comes from existing video grounding benchmarks. Our dataset includes a wide spectrum of video types from seven sources, including $(1)$ procedure videos \eg COIN~\citep{tang2019coin}, $(2)$ human activities \eg ActivityNet-Captions~\citep{yu2019activitynet} and Charades~\citep{krishna2017dense}, $(3)$ ego-centric videos \eg EgoExo4D~\citep{grauman2024ego}, $(4)$ movie descriptions~\citep{rohrbach15cvpr}, $(5)$ professional gymnasium videos \eg FineGym~\citep{shao2020finegym}, and $(6)$ unexpected humor videos Oops~\citep{oops}. We sample around 300 video clips from the validation and test sets of each video dataset, which results in 2K videos. The statistics of \shortname{} is shown in Table~\ref{tab:data_statistics}.

\begin{figure}[t]
    \centering
    \includegraphics[width=\linewidth]{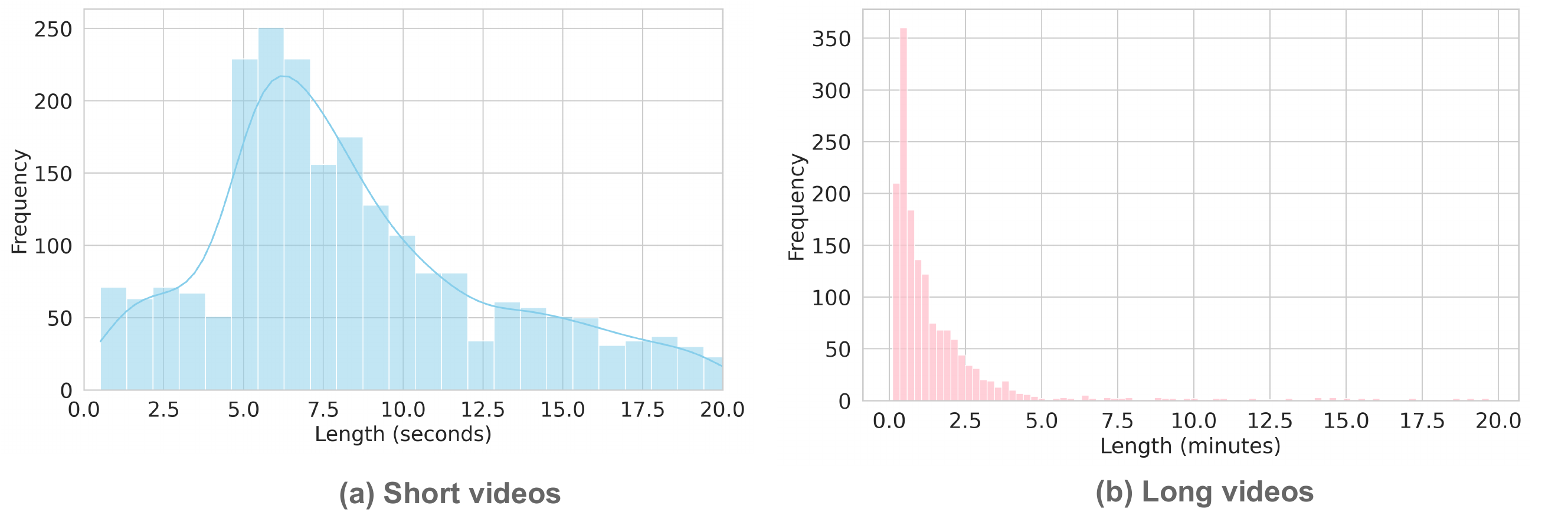}
\caption{Video length distribution of (a) short video clips and (b) long videos in \shortname{}. }
\label{fig:benchmark_video_clip_length_distribution}
\end{figure}

We intentionally filter out video clips that (1) are mostly static by leveraging optical flow~\citep{farneback2003two}, (2) contain multiple scene transitions by leveraging PySceneDetect~\footnote{\url{https://www.scenedetect.com/}} and (3) last longer than 20 seconds. We observe that the large amount of information in long videos make it difficult for annotators to provide detailed action descriptions. The distribution of video lengths is shown in Figure~\ref{fig:benchmark_video_clip_length_distribution} (a). Additionally, we remove the audio from the videos during annotation to ensure that all informative signals come solely from the visual frames, preventing the answers from being influenced by the audio.


\subsection{Video Caption  Annotation Process }

\textbf{Positive Captions Annotation.}
We employ a two-stage human labeling process for curating video captions with fine-grained activity descriptions, where the qualified Amazon Mechanical Turk (AMT) workers are first instructed to give a detailed video caption. Then, the authors of this work refine the caption by correcting the mistakes and adding missing details \textit{w.r.t.} the actions. The overall pipeline is shown in Figure~\ref{fig:benchmark_annotation}. All video clips are annotated following the same pipeline except for Finegym~\citep{shao2020finegym} as it has already provided  accurate and detailed action descriptions for professional gymnasium videos. Consequently, we reuse its annotations.

We first use 3 probing video captioning questions with 2 in-context examples as the onboarding task for  AMT master workers. We manually inspect the soundness and amount of temporal details of the AMT worker captions to select high quality AMT video captioning workers. During the annotation process by AMT workers, we also continue to remove the unqualified workers based on the ratio of the captions that authors in this paper refined. In this way, we ensure that the AMT provides a high quality initial point for positive captions. 

\textbf{Negative Caption Annotation.}
Our negative captions are aimed at confusing multimodal video models with respect to fine-grained activity details, such as changing \textit{``cut a ginger twice using a knife''} to \textit{``cut a ginger three times using a knife''}. 
We construct negatives upon two granularities: word level and event level. Specifically, word level negatives denote the case where  a certain word or phrase is replaced while event level negatives denote the case where the order of two events are reversed. Empirically, we find that LLMs can produce more creative and diverse negatives compared to AMT workers and authors. Therefore, we leverage three leading LLMs, GPT-4o~\citep{GPT4o}, Gemini-1.5-Pro~\citep{geminiteam2024gemini} and Llama-3.1-405b~\citep{llama-3} to curate a diverse set of negative caption candidates instructed by 3 in-context examples, with up to 9 negatives at word level and 6 negatives at event level. 

Afterwards, the authors of this work review those negative caption candidates in the format of multi-choice QA, which results in our complete \shortname{} dataset with $\sim$2K high-quality human-annotated video captions and $\sim$10K video question-answer pairs. 

\vspace{-5pt}

\subsection{A Pitfall in Multi-choice Question Answering}\label{sec:pitfall}

A conventional approach to evaluate large multimodal models is using the multi-choice question-answering format, which is adopted by the majority of current benchmarks including MMMU~\citep{yue2023mmmu}, MathVista~\citep{lu2024mathvistaevaluatingmathematicalreasoning}, EgoSchema~\citep{mangalam2023egoschema} etc. However, indicated by recent studies by~\citep{cai2024matryoshka} and~\citep{yue2024mmmupro}, a pure LLM can achieve comparable or even stronger performance on those benchmarks without looking at the visual content at all. Recent studies argue that (1) some questions are not designed well so that the question can be answered without looking at the visual content, or (2) the model memorizes the QA pairs, \ie data contamination occurs.

\begin{figure}[t]
    \centering
    \includegraphics[width=0.9\linewidth]{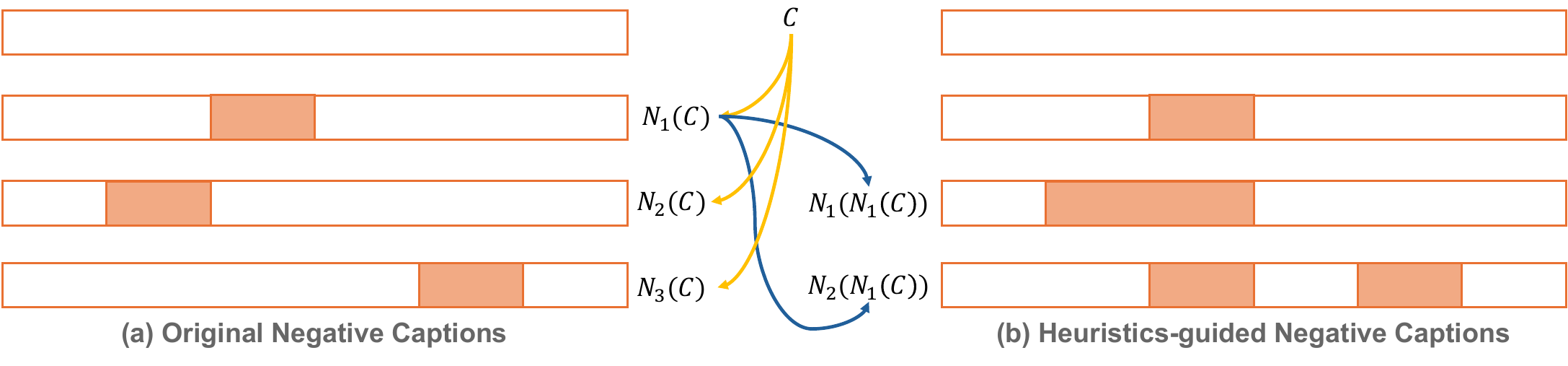}
    \vspace{-5pt}
    \caption{An illustration of multi-choice QA with (a) original and (b) heuristics-guided negative captions. Orange blocks indicate the altered contents from the positive option (green box).}
    \vspace{-15pt}
    \label{fig:negative_captions_generation}
\end{figure}

While developing our benchmark, we notice another previously ignored but critical pitfall for multi-choice QA. Specifically, if every negative answer choice is generated by changing a small part of the correct answer, the LLM can detect those changes to find a ``centralized" description and use that cue for its prediction.  To study this, given a positive caption $C$ and its associated negative caption $N(C)$, we intentionally derive a few negatives from $N_1(C)$ (instead of for $C$), resulting in  $N_1(N_1(C))$ and  $N_2(N_1(C))$, resulting in $[C, N_1(C), N_1(N_1(C)), N_2(N_1(C))]$ as options, so that $N_1(C)$ becomes the ``centralized'' description (see Fig.~\ref{fig:negative_captions_generation}). Surprisingly, we find that 66.4\% of text-only GPT-4o's predictions correspond to $N(C)$, while only 6.4\% of its predictions correspond to $C$. Our findings also align with human behavior analysis from psychology~\citep{furman2008similarity}, where humans can achieve better than random chance performance on multi-choice QAs using similar cues.

Motivated by this findings, we propose to decompose a single multi-choice QA into multiple binary QAs. In this case, we eliminate the ``centralized option" due to the fact that there are only two options to choose from. As a result, given $M$ negatives, the multiple binary QAs will query a model $M$ times, where the random chance performance changes from $\frac{1}{M+1}$ to $(\frac{1}{2})^M$. Given that $(\frac{1}{2})^M > \frac{1}{M+1}$ for every $M >2 $, multiple binary QA is a more difficult task than multi-choice QA.

\vspace{-5pt}
\section{Experiments}
\vspace{-5pt}

\subsection{Experiment Setup}

We evaluate both (1) multimodal video text generation models, including GPT-4o~\citep{GPT4o}, Gemini-1.5-Pro~\citep{geminiteam2024gemini}, Claude-3.5-Sonnet~\citep{claude}, Qwen2VL~\citep{Qwen2VL}, LLaVA-OneVision~\citep{li2024llavaonevision}, LLaVA-Next-Video~\citep{zhang2024llavanextvideo}, Phi-3.5-Vision~\citep{abdin2024phi}, MiniCPM-2.6~\citep{yao2024minicpm}, MA-LMM~\citep{he2024malmm}, VideoLLaVA~\citep{lin2023video}, InternLM-Xcomposer-2.5~\citep{internlmxcomposer2_5}, Matryoshka Multimodal Models~(\textit{$M^{3}$})~\citep{cai2024matryoshka}, 
and (2) multimodal video embedding models, including XCLIP~\citep{xclip}, ImageBind~\citep{girdhar2023imagebind}, and LanguageBind~\citep{zhu2023languagebind}. We exponentially increase the number of frames to study its effect on video understanding. More details can be found in Appendix~\ref{appendix: more results.}.

To study the effect of single frame bias and text bias, we also evaluate models trained on single images, including LLaVA-1.5~\citep{liu2023improvedllava}, LLaVA-NeXT~\citep{liu2024llavanext}, and Phi-3V~\citep{abdin2024phi}. In the latter case, we evaluate the LLMs including  GPT-4o~\citep{GPT4o}, Gemini-1.5-Pro~\citep{geminiteam2024gemini}, Yi-34B~\citep{yi}, Vicuna~\citep{vicuna2023} and Flan-T5~\citep{weifinetuned} without using videos at all.

\subsection{Human Performance}

We use Amazon Mechanical Turk to evaluate human performance. Note that we exclude the positive caption annotators to ensure that there is no data contamination. Again, we use an onboarding test  using a held out binary video QA evaluation set which has clear answers. Next, we show the performance on each task.

\subsection{Fine-grained Video Question Answering on Short Videos}\label{sec:fgvqa}

\begin{figure}[t]
    \centering
    \includegraphics[width=\linewidth]{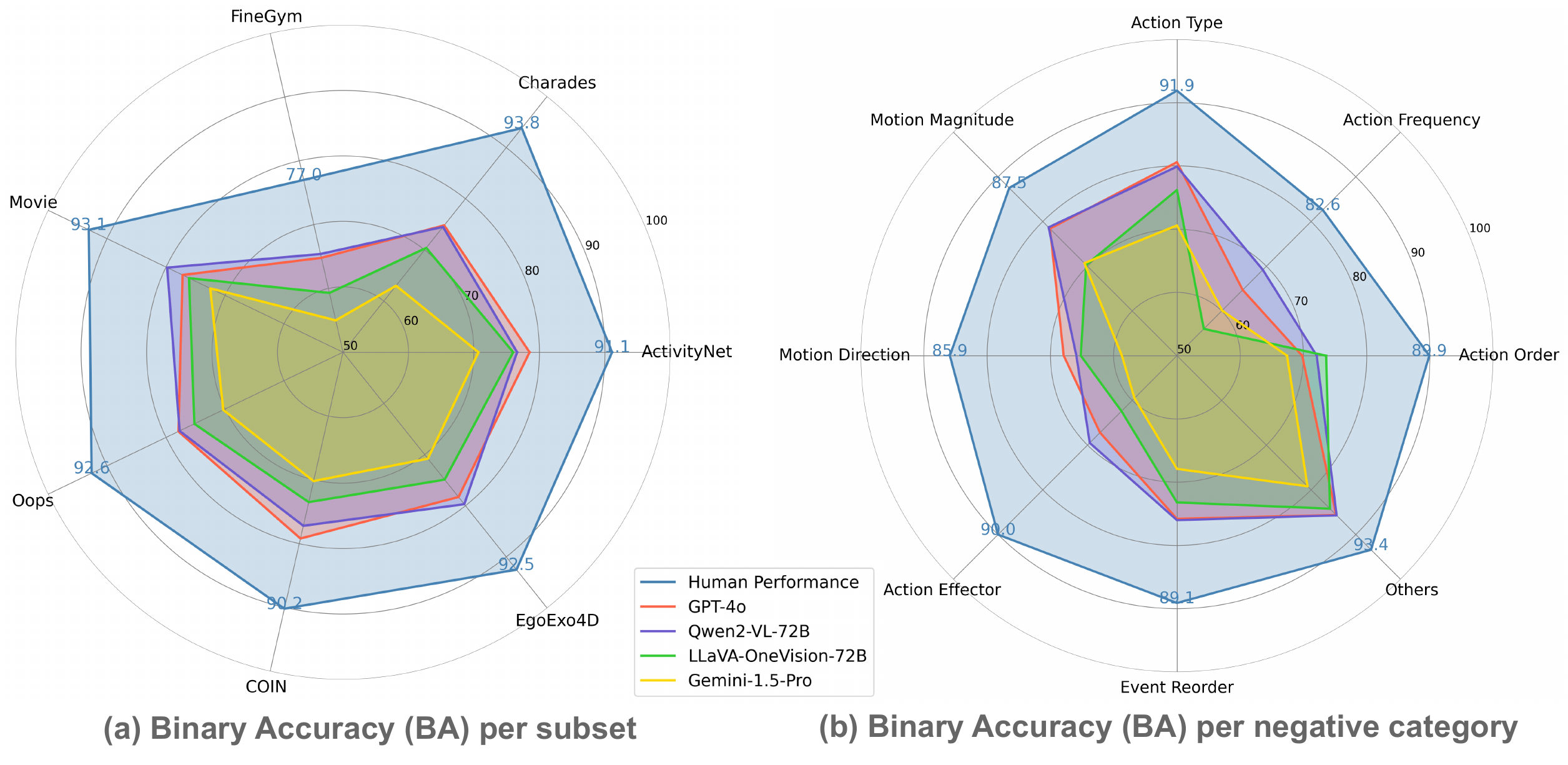}
    \vspace{-5pt}
    \caption{Visualization of binary accuracy for short video QA per (a) subset and (b) negative type. Human performance is much better than GPT-4o, Qwen2-VL-72B, LLaVA-OneVision-72B, and Gemini-1.5-Pro. }
    \label{fig:per_category_subset_performance}
\end{figure}

The results for multimodal generative models and embedding models are shown in Table~\ref{table:main_results} and Figure~\ref{fig:per_category_subset_performance} (a). Note that we show the result with the  best average multiple binary QA (MBA) performance for each model with respect to the number of frames. Results under different frames can be found in Appendix~\ref{appendix: more results.}. Several interesting findings arise:

\textbf{The performance of any video model is far from human performance.} As shown in the table, humans show an average performance of 67.9\%, which is significantly higher than the best models, GPT-4o and Qwen2VL-72B, by $\sim$30\%. Therefore, there is a large gap between model's performance and human performance. Note that we are employing standard AMT workers instead of domain experts, meaning that the expert-level accuracy can be even higher, especially for professional video understanding like FineGym.

\begin{figure}[t]
    \centering
    \includegraphics[width=0.7\linewidth]{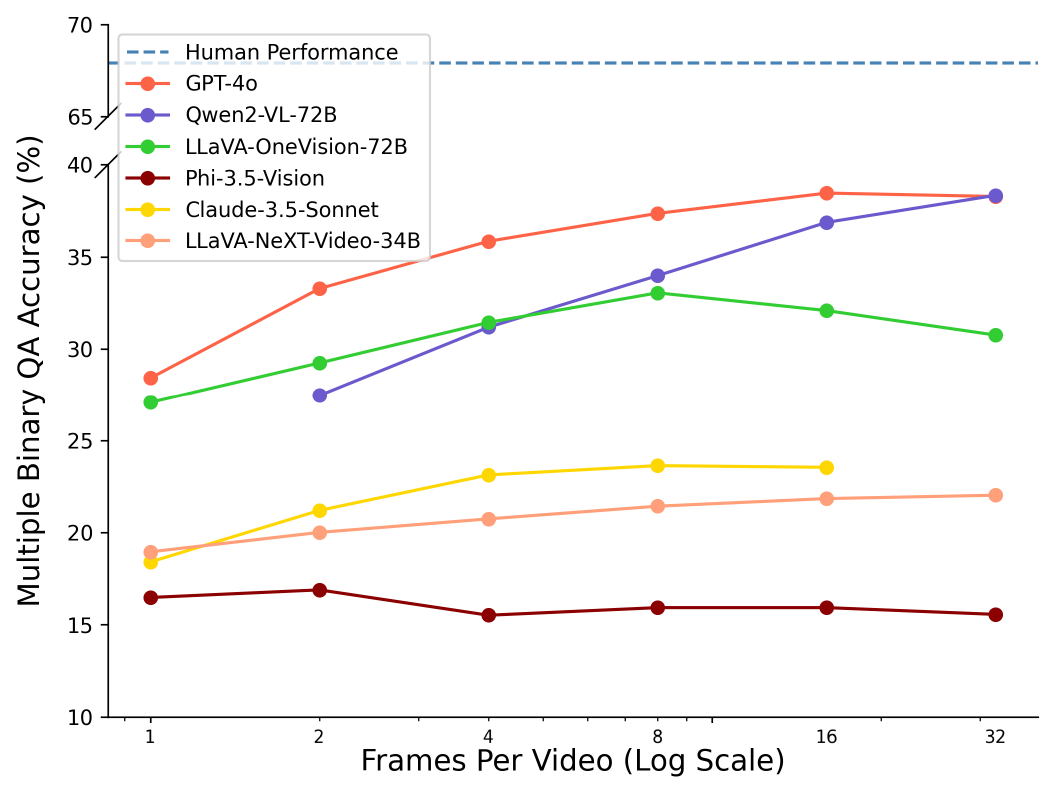}
    \caption{\textbf{Model performance on \shortname{}} with varying frames.}
    \label{fig:benchmark_video_performance_frames}
    \vspace{-10pt}
\end{figure}

\textbf{Models show limited performance gains with more frames}. As shown in Figure~\ref{fig:benchmark_video_performance_frames}, with more frames, multimodal video models usually show better performance. However, performance generally saturates around 8-16 frames, meaning that models struggle to improve fine-grained activity understanding even with more frames. This is a clear contrast with human performance, showing that there is still a large space for multimodal video models to improve.


\textbf{Multiple Binary QA is a more challenging metric. } Multiple Binary QA, as proposed in Section~\ref{sec:pitfall}, prevents a model from exploiting cues in the answer choices, and evaluates whether a model truly understands the temporal dynamics in the video by splitting a single $M+1$-way multiple choice question into $M$ binary choice questions.  For example, GPT-4o receives 75.7\% accuracy but only 38.5\% on multiple binary accuracy, showing a huge gap. These results indicate that understanding the fine-grained temporal dynamics is still a challenging task for current proprietary models and open-sourced models.

\textbf{Video Embedding models show near chance performance.} All multimodal video embedding models, including XCLIP, LanguageBind, and ImageBind show near random chance performance. One reason could be that their small embedding size (typically a vector with size around 768-2048) is insufficient to capture fine-grained temporal details.

\textbf{Low single-frame bias and language bias.}  As shown in Figure~\ref{fig:benchmark_video_performance_frames} and Table~\ref{tab:short_qa_full_table}, the performance of models like GPT-4o gradually increases with more frames. Excluding GPT-4o, all remaining VLMs trained with single images \eg LLaVA-1.5, Phi-3V, and text-only LLMs such as Yi-34B and Vicuna-7B show poor performance.

\begin{table}[t]
    \centering
    \caption{Dataset characteristics including number of samples, average number of words in original captions and our fine-grained captions. }
    \label{tab:data_statistics}
    \resizebox{\textwidth}{!}{
    \begin{tabular}{lccc}
    
        \toprule
        Dataset & Number of Samples & Org. Avg. \# words & Ours Avg. \# words \\
        \midrule
        ActivityNet~\citep{krishna2017dense} &  281 &13.03 &  49.55 \\
        EgoExo4D~\citep{grauman2024ego} & 307 & 7.73 & 47.79 \\
        Charades~\citep{gao2017tall} & 298 & 6.21 & 44.16 \\
        MPI Movie Description~\citep{rohrbach15cvpr} & 326 & 12.39 &35.33  \\
        Oops~\citep{oops} & 294 & 10.06 & 43.27 \\
        COIN~\citep{tang2019coin} & 385  &5.01  &  50.06\\
        \textcolor{gray}{ FineGym}~\citep{shao2020finegym} & 288 &21.92 & 21.92 \\ \midrule
        \shortname{} (ours) & 2179 & 10.91 & 41.72  \\
        \bottomrule
    \end{tabular}
    }
\end{table}

\begin{table}[t]
    \centering
    \caption{\shortname{} performance of various multimodal generative models and embedding models under the binary QA accuracy (BA) and multiple binary QA settings (MBA) for short videos. The prefix ``T-" indicates  MBA performance for the annotated subset in our \shortname{}. We show the result with the best average MBA performance for each model with respect to the number of frames, denoted as \# Frames.
    } 
    \label{table:main_results}
    \resizebox{\textwidth}{!}{
    \begin{tabular}{lc|ccccccc|cc}
        \toprule

Model              & \# Frames         & T-ActivityNet & T-Charades & T-FineGym & T-Movie & T-Oops & T-COIN & T-EgoExo4D & BA & MBA  \\\midrule
Human Performance     & -    & \textbf{68.7 }       & \textbf{82.2}     & \textbf{36.1 }   & \textbf{74.2  }& \textbf{69.7 }& \textbf{70.6 }& \textbf{71.0  } &\textbf{ 89.7} & \textbf{67.9 } \\
Random Chance     & -      &       11.0          & 13.7   &   6.1  & 12.0  & 5.6  &  11.1  & 5.6 &      50.0  &  9.5         \\ \midrule

\multicolumn{10}{c}{\textbf{Video Embedding Models: Text + Multiple Frames as Input}} \\ \midrule

XCLIP                 & 8    & 14.2        & 16.1     & 7.3     & 19.9  & 8.8  & 15.6 & 6.8    & 51.6 & 12.9  \\
ImageBind             & 2    & 17.4        & 16.8     & 7.3     & 19.0  & 11.2 & 16.1 & 9.1    & 53.0 & 14.0  \\
LanguageBind          & 8    & 22.4        & 15.1     & 6.6     & 19.3  & 10.9 & 15.6 & 11.1   & 52.8 & 14.5  \\

\midrule
\multicolumn{11}{c}{\textbf{Video Multimodal Generative Models : Text + Multiple Frames as Input}} \\ \midrule
GPT-4o                & 16   & \textbf{48.8}        & \textbf{42.6 }    & \textbf{18.8}    & 41.7  & 31.6 & \textbf{46.5} & 36.5   & 75.7 & \textbf{38.5}  \\
Gemini-1.5-Pro        & 1FPS & 34.9        & 24.5     & 8.3     & 35.6  & 22.8 & 34.3 & 21.8   & 67.5 & 26.6  \\
Claude-3.5-Sonnet     & 8    & 29.9        & 27.5     & 11.1    & 28.2  & 16.3 & 29.6 & 20.5   & 65.5 & 23.6  \\
Qwen2-VL-72B & 32   & 43.8        & 42.6     & 16.7    & \textbf{45.1}  & \textbf{36.7} & 43.6 & \textbf{37.1}   &\textbf{ 75.8} & 38.3  \\
Qwen2-VL-7B  & 32   & 32.4        & 32.2     & 4.9     & 35.9  & 18.4 & 25.5 & 21.8   & 64.4 & 24.7  \\
LLaVA-OneVision-72B   & 8    & 45.2        & 36.2     & 11.8    & 41.1  & 31.0 & 34.5 & 30.3   & 72.1 & 33.0  \\
LLaVA-OneVision-7B    & 32   & 30.2        & 23.2     & 5.9     & 27.3  & 18.0 & 25.5 & 16.3   & 61.9 & 21.2  \\
LLaVA-NeXT-Video-34B  & 32   & 30.6        & 26.8     & 10.4    & 24.8  & 18.0 & 25.2 & 17.3   & 64.0 & 22.0  \\
LLaVA-NeXT-Video-7B   & 8    & 33.5        & 32.6     & 10.8    & 28.2  & 17.3 & 22.9 & 19.9   & 65.1 & 23.6  \\
InternLM-XC2.5        & 1FPS & 25.3        & 21.5     & 8.7     & 24.8  & 11.9 & 18.4 & 14.0   & 58.8 & 17.9  \\
VideoLLaVA            & 8    & 35.2        & 29.2     & 13.5    & 25.5  & 20.7 & 32.5 & 20.2   & 67.1 & 25.5  \\
MiniCPM-V2.6          & 1FPS & 33.1        & 25.8     & 8.0     & 29.1  & 13.6 & 23.4 & 16.0   & 62.3 & 21.4  \\
Phi-3.5-Vision        & 2    & 25.3        & 20.1     & 5.2     & 22.7  & 12.2 & 18.2 & 13.7   & 58.0 & 16.9  \\
MA-LMM                & 4    & 12.5        & 16.4     & 3.5     & 11.0  & 5.1  & 11.4 & 4.9    & 48.0 & 9.4   \\
\textit{$M^{3}$}         & 6    & 21.0        & 20.1     & 6.6     & 19.6  & 10.2 & 15.1 & 10.4   & 56.4 & 14.8  \\

 \midrule
\multicolumn{11}{c}{\textbf{Large Multimodal Models (LMMs): Text + 1 Frame as Input}} \\ \midrule
GPT-4o                & 1    & 32.0        & 30.2     & 15.3    & 31.3  & 26.5 & 33.8 & 27.7   & 70.0 & 28.4  \\
LLaVA-1.5-13B         & 1    & 16.0        & 17.1     & 9.4     & 16.6  & 6.1  & 16.4 & 9.1    & 55.7 & 13.1  \\
LLaVA-1.5-7B          & 1    & 25.3        & 25.8     & 8.7     & 19.3  & 9.2  & 21.8 & 16.6   & 60.5 & 18.3  \\
LLaVA-NeXT-34B        & 1    & 20.6        & 22.5     & 9.4     & 21.5  & 15.3 & 21.6 & 13.7   & 60.5 & 18.0  \\
Phi-3-Vision          & 1    & 23.1        & 19.8     & 4.5     & 17.8  & 8.5  & 17.7 & 13.7   & 54.4 & 15.1  \\
  \midrule
\multicolumn{11}{c}{\textbf{Large Language Models (LLMs): Text as Input}} \\ \midrule
GPT-4o                & 0    & 30.2        & 31.9     & 16.7    & 27.9  & 22.8 & 27.5 & 28.0   & 67.7 & 26.5  \\
Gemini-1.5-Pro        & 0    & 22.4        & 20.5     & 4.5     & 19.9  & 10.2 & 16.9 & 17.9   & 58.1 & 16.1  \\
Yi-34B                & 0    & 17.4        & 27.5     & 10.4    & 21.8  & 11.2 & 23.4 & 16.9   & 59.9 & 18.7  \\
Vicuna7b-1-5          & 0    & 11.4        & 17.4     & 6.6     & 11.3  & 5.1  & 12.2 & 7.8    & 50.5 & 10.4  \\
Flan-T5-XL            & 0    & 24.9        & 23.5     & 5.6     & 19.9  & 11.9 & 23.4 & 14.0   & 57.9 & 17.9  \\
Flan-T5-XXL           & 0    & 19.2        & 16.8     & 8.3     & 18.1  & 7.8  & 19.7 & 14.0   & 55.1 & 15.1 
\\ \bottomrule
\end{tabular}
}
\end{table}

\subsection{Video Captioning}

Our detailed video captions also enables analyzing a model's fine-grained video captioning capabilities.
For this, we prompt multimodal video models to generate a caption for an input video, with 3 captioning examples in the prompt as guidance to mimic the style of our detailed video captions. Note that we remove the FineGym captions due to its different structure compared to other video captions, resulting in 1891 samples.  We evaluate the resulting video captioning performance using classical image captioning metrics,   CIDEr~\citep{vedantam2015cider}, BLEU~\citep{papineni-etal-2002-bleu} at different n-gram levels, ROUGE~\citep{lin-2004-rouge}, as well as the embedding similarity with sentence transformer~\citep{reimers-2019-sentence-bert} between the ground truth caption and the generated caption. Note that we for each model, we use the same number of frames as in Section~\ref{sec:fgvqa}.

Results in Table~\ref{tab:captioning} show that GPT-4o achieves the best performance. Interestingly, the results indicate that the embedding similarity aligns most closely with the video QA task results from Sec~\ref{sec:fgvqa}. Other classical captioning metrics show inconsistent results. For example, GPT-4o obtains similar performance with one compared to 64 frames on both CIDEr and BLEU scores (e.g., for BLEU\_1 24.1 vs. 25.1). On the other hand, all models show similar ROUGE scores. Thus, for the zero-shot captioning task, our findings indicate that text embedding similarity may be the most reliable metric.

\begin{table}[t!]
    \centering    
    \caption{Comparison of models for video captioning using Caption Similarity, CIDEr, BLEU, and ROUGE metrics. Cosine similarity using sentence transformer reflects the captioning quality the best.
    }
    \label{tab:captioning}
    \resizebox{\textwidth}{!}{
    
\begin{tabular}{llllllll}
\toprule
Model                                               & Similarity & CIDEr & ROUGE & BLEU\_1 & BLEU\_2 & BLEU\_3 & BLEU\_4  \\ \midrule
\multicolumn{8}{c}{\textbf{Video Multimodal Generative Models : Text + Multiple Frames as Input}} \\ \midrule
GPT-4o                & \textbf{61.3} & 7.3  & \textbf{19.6} & 24.1 & \textbf{11.8} & \textbf{5.8} & \textbf{3.0}  \\
Gemini-1.5-Pro        & 56.5 & \textbf{10.9} & 19.1 & 19.0 & 9.2  & 4.5 & 2.4  \\
Claude-3.5-Sonnet     & 54.1 & 8.6  & 17.1 & {24.4} & 10.3 & 4.4 & 2.1  \\
Qwen2-VL-72B & 56.1 & 9.3  & 19.1 & 15.7 & 8.0  & 4.1 & 2.2  \\
Qwen2-VL-7B  & 51.9 & 6.9  & 18.0 & 12.5 & 6.1  & 3.0 & 1.6  \\
LLaVA-OneVision-72B   & 55.0 & 9.7  & 18.7 & 23.7 & 11.3 & 5.6 & 2.9  \\
LLaVA-OneVision-7B    & 50.1 & 0.3  & 14.5 & 11.1 & 5.1  & 2.2 & 1.1  \\
LLaVA-NeXT-Video-34B  & 53.1 & 5.3  & 15.9 & 21.4 & 9.2  & 4.0 & 1.8  \\
LLaVA-NeXT-Video-7B   & 50.1 & 2.3  & 15.8 & 18.1 & 7.0  & 2.6 & 1.1  \\
InternLM-XC2.5        & 52.4 & 2.3  & 15.9 & 17.8 & 7.1  & 2.8 & 1.2  \\
VideoLLaVA            & 46.0 & 4.5  & 16.9 & 12.6 & 5.4  & 2.3 & 1.0  \\
MiniCPM-V2.6          & 47.2 & 1.5  & 14.2 & 15.5 & 5.4  & 1.9 & 0.8  \\
Phi-3.5-Vision        & 42.9 & 3.7  & 16.5 & 20.4 & 8.4  & 3.4 & 1.6  \\
MA-LMM                & 38.7 & 3.1  & 15.0 & 10.1 & 4.8  & 2.2 & 1.1  \\
\textit{$M^{3}$}         & 47.8 & 3.0  & 16.4 & 16.7 & 6.9  & 2.8 & 1.2  \\
    \midrule
\multicolumn{8}{c}{\textbf{Large Multimodal Models (LMMs): Text + 1 Frame as Input}} \\ \midrule
GPT-4o                & 52.3 & 7.3  & 17.1 & \textbf{25.1} & 11.1 & 5.0 & 2.4  \\
LLaVA-1.5-13B         & 47.9 & 4.9  & 18.0 & 22.6 & 9.8  & 4.2 & 2.0  \\
LLaVA-1.5-7B          & 45.7 & 6.9  & 17.8 & 22.0 & 9.5  & 4.2 & 2.0  \\
LLaVA-NeXT-34B        & 49.1 & 6.2  & 16.7 & 24.2 & 10.4 & 4.6 & 2.2  \\
Phi-3-Vision          & 42.0 & 4.0  & 16.1 & 19.9 & 8.3  & 3.4 & 1.6  \\ \bottomrule
\end{tabular}
}
\end{table}

\begin{table}[t!]
    \centering
    \caption{\shortname{} performance of various multimodal generative models and embedding models under \textbf{long video} understanding with binary QA accuracy (BA) and multiple binary QA accuracy (MBA). The MBA performance under each dataset is also included. We show the result with the best average MBA performance for each model with respect to the number of frames, denoted as \# Frames. }
    \label{table:longvideo}
    \resizebox{\textwidth}{!}{
    \begin{tabular}{lc|ccccc|cc}
        \toprule

Model                      & \# Frames      & T-ActivityNet & T-Charades & T-FineGym & T-COIN & T-EgoExo4D  & BA &  MBA \\\midrule
Random Performance   & -    & 9.2         & 4.3      & 11.2    & 11.4 & 9.9    & 50.2            & 9.5                       \\
\midrule
\multicolumn{9}{c}{\textbf{Video Embedding Models: Text + Multi-Frames as Input}} \\ \midrule
XCLIP                & 8    & 11.1        & 12.4     & 6.5     & 10.8 & 11.8   & 51.7            & 11.1                      \\
ImageBind            & 2    & 10.2        & 8.1      & 9.3     & 10.8 & 12.4   & 51.0            & 10.7                      \\
LanguageBind         & 8    & 11.7        & 10.8     & 10.3    & 11.0 & 14.1   & 51.6            & 12.0                      \\
 \midrule

\multicolumn{9}{c}{\textbf{Video Multimodal Generative Models : Text + Multi-Frames as Input}} \\ \midrule

GPT-4o               & 64   & \textbf{40.0}        & \textbf{37.8}     & 16.8    &\textbf{ 32.7 }& 29.3   & \textbf{70.5}            & \textbf{32.7  }                    \\
Gemini-1.5-Pro       & 1FPS & 32.1        & 18.4     & 18.7    & 24.8 & 23.8   & 65.2            & 24.7                      \\
Claude-3.5-Sonnet    & 8    & 28.9        & 22.2     & 16.8    & 22.2 & 26.7   & 64.6            & 24.5                      \\
Qwen2-VL-72B         & 8    & 32.4        & 20.5     & 21.5    & 18.9 & 33.1   & 64.7            & 26.2                      \\
Qwen2-VL-7B          & 32   & 22.2        & 20.0     & 9.3     & 18.3 & 18.7   & 59.7            & 18.8                      \\
LLaVA-OneVision-72B  & 4    & 28.6        & 19.5     & 18.7    & 16.5 & 30.9   & 63.4            & 23.8                      \\
LLaVA-OneVision-7B   & 32   & 21.3        & 13.0     & 13.1    & 11.4 & 19.8   & 56.9            & 16.2                      \\
LLaVA-NeXT-Video-34B & 4    & 23.5        & 22.2     & 19.6    & 17.9 & 19.2   & 60.3            & 20.0                      \\
LLaVA-NeXT-Video-7B  & 8    & 18.1        & 21.6     & 10.3    & 18.5 & 15.6   & 57.2            & 17.3                      \\
InternLM-XC2.5       & 1FPS & 21.0        & 18.4     & 20.6    & 14.0 & 11.4   & 55.8            & 15.6                      \\
VideoLLaVA           & 8    & 20.0        & 16.8     & 15.9    & 9.8  & 16.6   & 56.0            & 15.1                      \\
MiniCPM-V2.6         & 1FPS & 14.3        & 16.8     & 6.5     & 17.1 & 14.1   & 60.3            & 19.3                      \\
Phi-3.5-Vision       & 4    & 23.2        & 11.9     & 19.6    & 10.2 & 13.3   & 54.5            & 14.5                      \\
MA-LMM               & 4    & 10.2        & 9.2      & 2.8     & 11.4 & 11.6   & 47.1            & 9.2                       \\
\textit{$M^{3}$}      & 6    & 10.8        & 8.6      & 12.1    & 13.0 & 12.4   & 53.1            & 11.8                      \\


 \midrule
\multicolumn{9}{c}{\textbf{Large Multimodal Models (LMMs): Text + 1 frame as Input}} \\ \midrule

GPT-4o               & 1    & 27.9        & 23.2     & 19.6    & 25.2 & 22.9   & 64.7            & 24.5                      \\
LLaVA-1.5-13B        & 1    & 14.3	  & 11.9	  & 10.3  & 	15.4	  & 14.7	  & 54.8  & 14.2                \\
LLaVA-1.5-7B         & 1    & 9.2         & 11.9     & 10.3    & 12.8 & 14.5   & 53.2            & 12.3                      \\
LLaVA-NeXT-34B       & 1    & 21.6        & 20.5     & 19.6    & 18.9 & 19.8   & 60.5            & 19.9                      \\
Phi-3-Vision         & 1    & 18.1        & 12.4     & 15.0    & 15.4 & 15.6   & 56.0            & 15.6                      \\

\midrule
\multicolumn{9}{c}{\textbf{Large Larguage Models (LLMs): Text as Input}} \\ \midrule
GPT-4o               & 0    & 27.6        & 32.4     & 17.8    & 24.2 & \textbf{33.5 }  & 67.6            & 28.2                      \\
Gemini-1.5-Pro       & 0    & 22.9        & 19.5     & 17.8    & 19.3 & 23.4   & 62.2            & 21.2                      \\
Yi-34B               & 0    & 19.7        & 19.5     & 14.0    & 15.9 & 20.6   & 59.5            & 18.4                      \\
Vicuna7b-1-5         & 0    & 6.3         & 9.2      & 9.3     & 10.6 & 12.0   & 51.1            & 9.9                       \\
Flan-T5-XL           & 0    & 21.6        & 15.7     & \textbf{23.4}    & 18.1 & 19.8   & 60.1            & 19.4                      \\
Flan-T5-XXL          & 0    & 20.0        & 11.9     & 18.7    & 15.7 & 17.1   & 56.9            & 16.7

\\ \bottomrule
\end{tabular}
}
\end{table}

\subsection{Long Video Understanding}

Since our benchmark is annotated at the video clip level, we can easily extend it to long video understanding by concatenating the captions of different video clips within the same original video. 
In our study, we choose video datasets from AcitivityNet, Charades, EgoExo4D, COIN and FineGym. 
We randomly sample video clips within the same original video, and then crop a new video segment whose starting time corresponds to that of the earliest sampled video clip and whose ending time corresponds to that of the latest sampled video clip.  We then concatenate all the sampled video captions together to form a single long detailed description corresponding to the new video segment. Given this positive caption, we generate negative captions for it by replacing the positive caption of one of the sampled video clips with its negatives. The model is then tasked to choose the correct long caption out of multiple choices.  We control the random chance multiple binary QA performance to be $\sim$9.5\%, resulting in an apple-to-apple comparsion with in Sec~\ref{sec:fgvqa}.  In this way, we investigate whether multimodal video models can understand and distinguish fine details in a long video. Finally, we sampled 1,574 videos with durations ranging between $[0, 20]$ minutes, as shown in Figure~\ref{fig:benchmark_video_clip_length_distribution}.

We show in Table~\ref{table:longvideo}, that all multimodal video models show a significant performance drop for this task compared to short video understanding. This is also reflected in all models performing better on relatively shorter videos (\eg Charades) compared to longer videos (\eg  FineGym). These results indicate that finding the subtle temporal dynamic differences in a long video is indeed an extremely difficult task.  It is similar in nature to the needle-in-the-sea task~\citep{kam2023} in NLP except in the temporal domain. We hope that \shortname{} for long video understanding can serve as a very challenging task for future video understanding model development.

\begin{table}[h]
    \centering
    \caption{Effect of the ``Centralized" Caption on text-only GPT-4o. }
\label{tab:centralizaed analysis}
    \resizebox{0.55\textwidth}{!}{
    \begin{tabular}{ccc}
        \toprule
Percentage of Predictions Aligned with $\longrightarrow$ & $C$   & $N_1(C)$ \\ \midrule
$[C, N_1(C), N_2(C)), N_3(C))]$                    & \textbf{83.3} & 6.4  \\
$[C, N_1(C), N_1(N_1(C)), N_2(N_1(C))]$                   & 17.7 & \textbf{66.4}    \\ \bottomrule
\end{tabular}
}
\vspace{-1em}
\end{table}


\section{In-Depth Analysis}

\subsection{Why multiple Binary QA instead of multi-choice QA?}

As discussed in Section~\ref{sec:pitfall}, in the standard multi-choice QA setting, if negatives are all slightly variations of the positive caption, we find that LLMs can determine the ``centralized" caption, and take a shortcut to achieve better performance. To demonstrate this, based on one negative caption $N(C)$ in \shortname{}, we intentionally generate two negative captions derived from  $N(C)$ (instead of $C$), resulting in $N_1(N(C))$ and $N_2(N(C))$. Given two set of options $[C, N_1(C), N_2(C)), N_3(C))]$  and $[C, N_1(C), N_1(N_1(C)), N_2(N_1(C))]$ shown in Figure~\ref{fig:negative_captions_generation}, text-only GPT-4o displays different behaviors. As shown in Table~\ref{tab:centralizaed analysis}, under the intentionally designed negative options, GPT-4o will choose $N_1(C)$  under 66.4\%  cases. This again demonstrates the necessity and advantage of our multiple binary QA accuracy (MBA) metric design over the standard multi-choice QA setting.


\subsection{Performance on categories}

\begin{figure}[t]
    \centering
    \includegraphics[width=\linewidth]{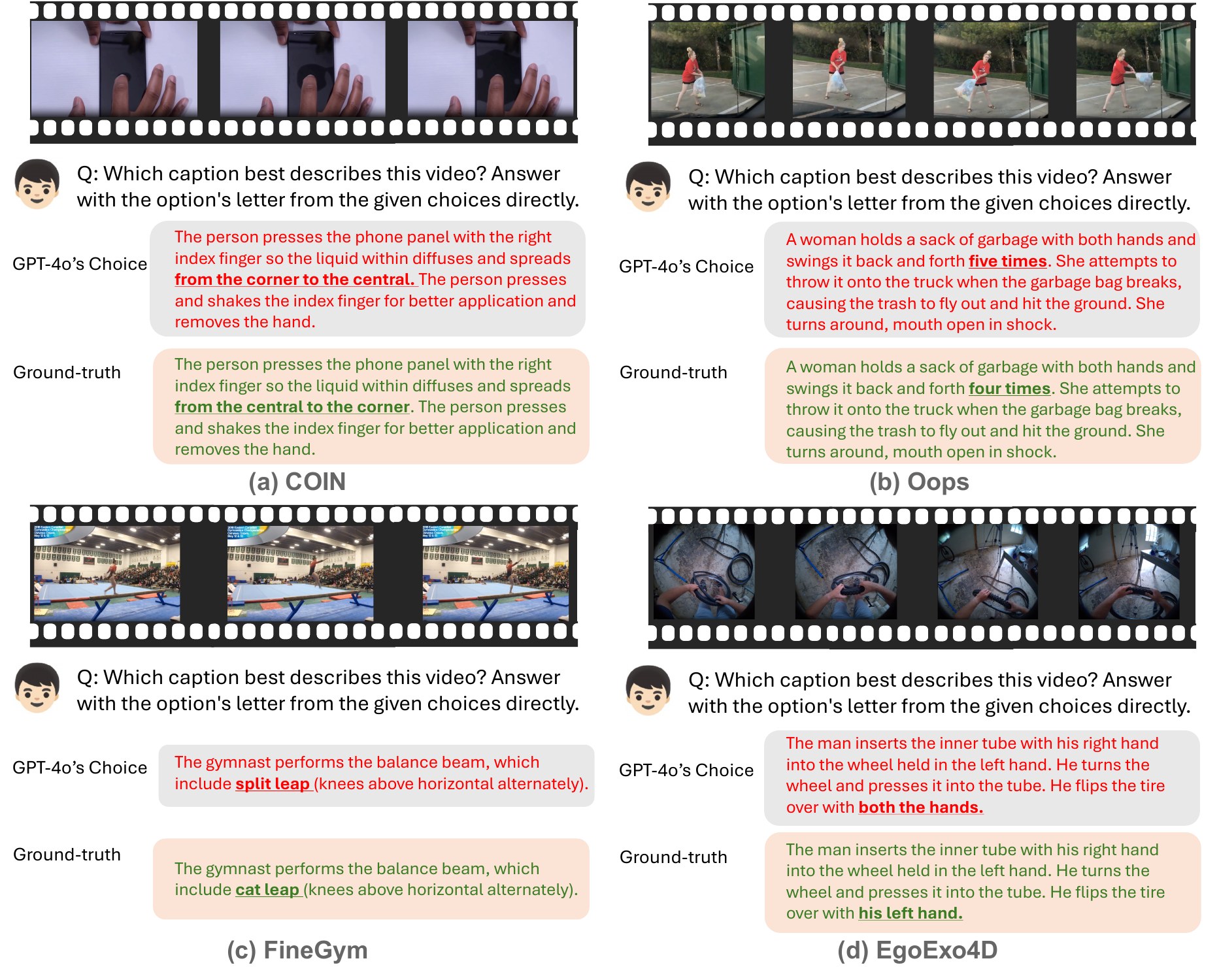}
    \caption{\textbf{The failure cases of GPT-4o in \shortname{}}. GPT-4o does not understand the fine-grained details well, including motion direction, action frequency, action type, and motion direction.}
    \label{fig:gpt4o failure analysis}
\end{figure}

Broadly, \shortname{} evaluates word level replacement and event level re-ordering. Here we further breakdown the word level replacement into following categories: $(1).$ Action order (change the order);  $(2).$ Action frequency (1 times \textit{v.s.} two times); $(3).$ Action type (put \textit{v.s.} pull); $(4).$ Motion magnitude (slightly \textit{v.s.} intensively); $(5).$ Motion Direction/Orientation (forward \textit{v.s.} backward, circular \textit{v.s.} back-and-forth). $(6).$ Action effector (cutting with left hand \textit{v.s.} cutting with right hand) $(7).$ Others. We prompt GPT-4o to perform 7-way classification and show the per-category performance in Table~\ref{table:category_results}  and Figure~\ref{fig:per_category_subset_performance} (b). Results indicate that  multimodal video models shows better performance on ``others" category rather than the other categories related to actions. Among the seven categories, models struggle most on action frequency (counting), which show that they do not memorize repeated occurrences well. The visualizations of failture cases in GPT-4o is shown in Figure~\ref{fig:gpt4o failure analysis}.

\begin{table}
\centering
\caption{\shortname{} statistics on negative caption types.}
\label{table:category_breakdown}
 \resizebox{0.8\textwidth}{!}{
\begin{tabular}{@{}lcccccccr@{}}
\toprule
{Action} & {Action} & {Action} & {Motion} & {Motion} & {Action} & {Event} & {} & {Overall} \\
{Order} & {Frequency} & {Type} & {Magnitude} & {Direction} & {Effector} & {Reorder} & {Others} & \\
\midrule
129 & 530 & 2,802 & 320 & 1,536 & 1,109 & 2,099 & 1,342 & 9,867 \\
\bottomrule
\end{tabular}
}
\end{table}

\begin{table}[t]
    \centering
    \caption{\shortname{} performance under each category under BA. Multimodal videos models struggle on certain tasks such as action frequency. We show the result with the  best average MBA performance for each model with respect to the number of frames.}
    \label{table:category_results}
    \resizebox{\textwidth}{!}{
    \begin{tabular}{lc|cccccccc|c}
        \toprule

                      & The Number & Action & Action & Action & Motion & Motion & Action & Event & &  \\
Model                  & of Frames    & Order & frequency & Type & Magnitude & Direction & Effector & Reorder & Others & Average   
\\\midrule
Human Performance     & -    &\textbf{ 89.9} & \textbf{82.6} & \textbf{91.9} & \textbf{87.5} & \textbf{85.9} &\textbf{ 90.0 }& \textbf{89.1 }& \textbf{93.4}   & \textbf{89.7}  \\
Random Chance         & -    & 50.0  & 50.0 & 50.0 & 50.0 & 50.0 & 50.0 & 50.0 & 50.0 & 50.0 \\\midrule
\multicolumn{10}{c}{\textbf{Video Embedding Models: Text + Multi-Frames as Input}} \\ \midrule
XCLIP                 & 8    & 46.5 & 50.8 & 50.9 & 56.9 & 51.2 & 51.7 & 50.1 & 55.6   & 51.6  \\
ImageBind             & 2    & 44.2 & 44.7 & 55.4 & 50.9 & 52.5 & 50.5 & 48.6 & 61.8   & 53.0  \\
LanguageBind          & 8    & 43.4 & 41.5 & 53.4 & 55.0 & 51.4 & 46.6 & 51.0 & 65.9   & 52.8  \\
  \midrule
\multicolumn{10}{c}{\textbf{Video Multimodal Generative Models : Text + Multi-Frames as Input}} \\ \midrule
GPT-4o                & 16   & 69.8 & 64.7 & \textbf{80.6} &{ 78.4 }&\textbf{ 67.9} & 67.2 & 75.8 & 85.6   & 75.7  \\
Gemini-1.5-Pro        & 1FPS & 67.4 & 60.1 & 70.6 & 70.7 & 58.7 & 59.5 & 67.9 & 79.2   & 67.5  \\
Claude-3.5-Sonnet     & 8    & 62.0 & 57.4 & 70.7 & 70.3 & 60.0 & 57.8 & 61.3 & 76.2   & 65.5  \\
Qwen2-VL-72B & 32   & 72.1 & 69.2 & 79.9 & \textbf{78.7} & 65.9 & 69.5 & \textbf{76.0} & \textbf{85.7 }  & \textbf{75.8}  \\
Qwen2-VL-7B  & 32   & 65.9 & 45.8 & 67.3 & 66.1 & 54.6 & 54.7 & 69.7 & 75.7   & 64.4  \\
LLaVA-OneVision-72B   & 8    & \textbf{73.6} & 56.0 & 76.2 & 70.3 & 65.2 & 62.4 & 73.2 & 84.2   & 72.1  \\
LLaVA-OneVision-7B    & 32   & 63.6 & 45.5 & 62.9 & 56.9 & 52.8 & 54.0 & 66.5 & 77.1   & 61.9  \\
LLaVA-NeXT-Video-34B  & 32   & 61.2 & 56.0 & 66.4 & 61.6 & 58.5 & 59.3 & 63.4 & 74.1   & 64.0  \\
LLaVA-NeXT-Video-7B   & 8    & 69.0 & 65.7 & 68.2 & 62.2 & 66.5 & 68.6 & 52.2 & 74.3   & 65.1  \\
InternLM-XC2.5        & 1FPS & 55.8 & 42.5 & 62.7 & 62.5 & 52.6 & 51.1 & 58.3 & 70.7   & 58.8  \\
VideoLLaVA            & 8    & 69.8 &\textbf{ 70.2} & 71.4 & 70.0 & 70.6 & \textbf{70.2} & 50.5 & 75.5   & 67.1  \\
MiniCPM-V2.6          & 1FPS   & 59.4 & 52.3 & 65.5 & 62.5 & 54.1 & 53.3 & 63.5 & 74.7   & 62.3  \\
Phi-3.5-Vision        & 2    & 53.5 & 55.3 & 60.1 & 55.9 & 54.0 & 52.2 & 55.3 & 69.4   & 58.0  \\
MA-LMM                & 4    & 54.3 & 43.0 & 48.0 & 47.8 & 46.3 & 48.8 & 48.6 & 49.6   & 48.0  \\
\textit{$M^{3}$}        & 6    & 51.9 & 53.6 & 58.9 & 56.3 & 52.2 & 53.7 & 50.8 & 68.6   & 56.4  \\

\midrule
\multicolumn{10}{c}{\textbf{Large Multimodal Models (LMMs): Text + 1 frame as Input}} \\ \midrule
GPT-4o                & 1    & 67.4 & 65.1 & 74.1 & 70.3 & 64.2 & 62.6 & 68.7 & 78.4   & 70.0  \\
LLaVA-1.5-13B         & 1    & 57.4 & 51.9 & 57.6 & 53.8 & 50.4 & 53.9 & 54.2 & 63.1   & 55.7  \\
LLaVA-1.5-7B          & 1    & 62.0 & 61.5 & 62.2 & 54.1 & 61.4 & 64.9 & 51.0 & 67.9   & 60.5  \\
LLaVA-NeXT-34B        & 1    & 51.2 & 55.7 & 61.2 & 60.0 & 54.8 & 53.0 & 65.0 & 67.5   & 60.5  \\
Phi-3-Vision          & 1    & 46.5 & 45.5 & 56.0 & 55.6 & 48.8 & 49.2 & 56.9 & 62.1   & 54.4  \\
 \midrule
\multicolumn{10}{c}{\textbf{Large Larguage Models (LLMs): Text as Input}} \\ \midrule
GPT-4o                & 0    & 65.1 & 59.8 & 73.7 & 70.0 & 61.5 & 60.1 & 69.3 & 68.6   & 67.7  \\
Gemini-1.5-Pro        & 0    & 54.3 & 42.5 & 60.4 & 62.2 & 53.6 & 53.3 & 64.8 & 57.4   & 58.1  \\
Yi-34B                & 0    & 51.9 & 62.3 & 60.1 & 60.3 & 57.1 & 55.1 & 65.4 & 58.0   & 59.9  \\
Vicuna7b-1-5          & 0    & 55.8 & 47.2 & 51.7 & 48.4 & 50.1 & 49.4 & 49.9 & 51.4   & 50.5  \\
Flan-T5-XL            & 0    & 53.5 & 57.7 & 60.2 & 59.7 & 56.1 & 56.9 & 54.9 & 60.7   & 57.9  \\
Flan-T5-XXL           & 0    & 55.8 & 62.5 & 59.0 & 58.4 & 54.2 & 48.2 & 49.3 & 58.9   & 55.1 
\\ \bottomrule
\end{tabular}
}
\end{table}

\section{Conclusion and Future Work}
\label{sec:conclusion and limitation}

We propose TemporalBench, a novel video understanding benchmark, to  evaluate the fine-grained temporal understanding abilities of multimodal video models. The video captions in our benchmark are significantly denser than existing datasets such as MSRVTT and TGIF, offering detailed temporal annotations.  TemporalBench also provides a more challenging set of tasks that push current multimodal models beyond coarse-level understanding. The empirical results reveal a substantial gap between human performance and current state-of-the-art models. We also found a critical pitfall for multi-
choice QA, where we devise multiple binary accuracy (MBA) to address thi issue.  We hope that this benchmark fosters further research in developing models with enhanced temporal reasoning capabilities. Our benchmark could also be easily utilized for other fundamental video tasks such as spatio-temporal localization and text-to-video generation with fine-grained prompts.

\textbf{Limitations.}
One cannot fully analyze the behavior of proprietary models included in this paper due to the lack of access to these models, which are GPT-4o, Gemini-1.5-Pro and Claude 3.5 Sonnet.

\section*{Acknowledgement}
This work was supported in part by NSF IIS2404180, and Institute of Information \& communications Technology Planning \& Evaluation(IITP) grants funded by the Korea government(MSIT) (No. 2022-0-00871, Development of AI Autonomy and Knowledge Enhancement for AI Agent Collaboration) and (No. RS2022-00187238, Development of Large Korean Language Model Technology for Efficient Pre-training), and Microsoft Accelerate Foundation Models Research Program.

\section*{Reproducibility Statement}
We attach part of the dataset in the submission's supplementary materials. We will also publicly release it along with the code used to evaluate the LMMs upon the paper's acceptance.

\section*{Ethics Statement}  
This research primarily utilizes publicly available video datasets, which have been collected and annotated by qualified annotators and authors, ensuring compliance with ethical standards. We have made every effort to ensure that the data used respects privacy and contains no personally identifiable information. Furthermore, we acknowledge the potential implications of fine-grained video understanding, especially in sensitive applications such as surveillance and autonomous systems. As such, we advocate for responsible and ethical use of this research, urging caution in deploying these models in real-world scenarios to avoid harmful or unintended consequences.

\bibliography{iclr2025_conference}

\begin{thebibliography}{87}
\providecommand{\natexlab}[1]{#1}
\providecommand{\url}[1]{\texttt{#1}}
\expandafter\ifx\csname urlstyle\endcsname\relax
  \providecommand{\doi}[1]{doi: #1}\else
  \providecommand{\doi}{doi: \begingroup \urlstyle{rm}\Url}\fi

\bibitem[Abdin et~al.(2024)Abdin, Jacobs, Awan, Aneja, Awadallah, Awadalla, Bach, Bahree, Bakhtiari, Behl, et~al.]{abdin2024phi}
Marah Abdin, Sam~Ade Jacobs, Ammar~Ahmad Awan, Jyoti Aneja, Ahmed Awadallah, Hany Awadalla, Nguyen Bach, Amit Bahree, Arash Bakhtiari, Harkirat Behl, et~al.
\newblock Phi-3 technical report: A highly capable language model locally on your phone.
\newblock \emph{arXiv preprint arXiv:2404.14219}, 2024.

\bibitem[Anthropic(2024)]{claude}
Anthropic.
\newblock Claude-sonnet-3.5.
\newblock \url{https://www.anthropic.com/news/claude-3-5-sonnet}, 2024.

\bibitem[Antol et~al.(2015)Antol, Agrawal, Lu, Mitchell, Batra, Zitnick, and Parikh]{antol2015vqav1}
Stanislaw Antol, Aishwarya Agrawal, Jiasen Lu, Margaret Mitchell, Dhruv Batra, C~Lawrence Zitnick, and Devi Parikh.
\newblock Vqa: Visual question answering.
\newblock In \emph{Proceedings of the IEEE international conference on computer vision}, pp.\  2425--2433, 2015.

\bibitem[Bai et~al.(2023)Bai, Bai, Yang, Wang, Tan, Wang, Lin, Zhou, and Zhou]{Qwen-VL}
Jinze Bai, Shuai Bai, Shusheng Yang, Shijie Wang, Sinan Tan, Peng Wang, Junyang Lin, Chang Zhou, and Jingren Zhou.
\newblock Qwen-vl: A versatile vision-language model for understanding, localization, text reading, and beyond.
\newblock \emph{arXiv preprint arXiv:2308.12966}, 2023.

\bibitem[Cai et~al.(2024{\natexlab{a}})Cai, Liu, Mustikovela, Meyer, Chai, Park, and Lee]{cai2024vipllava}
Mu~Cai, Haotian Liu, Siva~Karthik Mustikovela, Gregory~P. Meyer, Yuning Chai, Dennis Park, and Yong~Jae Lee.
\newblock Making large multimodal models understand arbitrary visual prompts.
\newblock In \emph{IEEE Conference on Computer Vision and Pattern Recognition}, 2024{\natexlab{a}}.

\bibitem[Cai et~al.(2024{\natexlab{b}})Cai, Yang, Gao, and Lee]{cai2024matryoshka}
Mu~Cai, Jianwei Yang, Jianfeng Gao, and Yong~Jae Lee.
\newblock Matryoshka multimodal models.
\newblock \emph{arXiv preprint arXiv:2405.17430}, 2024{\natexlab{b}}.

\bibitem[Chen \& Dolan(2011)Chen and Dolan]{chen2011msvd}
David~L. Chen and William~B. Dolan.
\newblock Collecting highly parallel data for paraphrase evaluation.
\newblock In \emph{Proceedings of the 49th Annual Meeting of the Association for Computational Linguistics (ACL-2011)}, Portland, OR, June 2011.

\bibitem[Chen et~al.(2023)Chen, Zhang, Zeng, Zhang, Zhu, and Zhao]{chen2023shikra}
Keqin Chen, Zhao Zhang, Weili Zeng, Richong Zhang, Feng Zhu, and Rui Zhao.
\newblock Shikra: Unleashing multimodal llm's referential dialogue magic.
\newblock \emph{arXiv preprint arXiv:2306.15195}, 2023.

\bibitem[Chen et~al.(2024)Chen, Wei, Li, Dong, Zhang, Zang, Chen, Duan, Lin, Tang, et~al.]{chen2024sharegpt4video}
Lin Chen, Xilin Wei, Jinsong Li, Xiaoyi Dong, Pan Zhang, Yuhang Zang, Zehui Chen, Haodong Duan, Bin Lin, Zhenyu Tang, et~al.
\newblock Sharegpt4video: Improving video understanding and generation with better captions.
\newblock \emph{arXiv preprint arXiv:2406.04325}, 2024.

\bibitem[Chiang et~al.(2023)Chiang, Li, Lin, Sheng, Wu, Zhang, Zheng, Zhuang, Zhuang, Gonzalez, Stoica, and Xing]{vicuna2023}
Wei-Lin Chiang, Zhuohan Li, Zi~Lin, Ying Sheng, Zhanghao Wu, Hao Zhang, Lianmin Zheng, Siyuan Zhuang, Yonghao Zhuang, Joseph~E. Gonzalez, Ion Stoica, and Eric~P. Xing.
\newblock Vicuna: An open-source chatbot impressing gpt-4 with 90\%* chatgpt quality, March 2023.
\newblock URL \url{https://lmsys.org/blog/2023-03-30-vicuna/}.

\bibitem[Chung et~al.(2024)Chung, Hou, Longpre, Zoph, Tay, Fedus, Li, Wang, Dehghani, Brahma, et~al.]{chung2022scalinginstructionfinetunedlanguagemodels}
Hyung~Won Chung, Le~Hou, Shayne Longpre, Barret Zoph, Yi~Tay, William Fedus, Yunxuan Li, Xuezhi Wang, Mostafa Dehghani, Siddhartha Brahma, et~al.
\newblock Scaling instruction-finetuned language models.
\newblock \emph{Journal of Machine Learning Research}, 25\penalty0 (70):\penalty0 1--53, 2024.

\bibitem[Epstein et~al.(2020)Epstein, Chen, and Vondrick]{oops}
Dave Epstein, Boyuan Chen, and Carl Vondrick.
\newblock Oops! predicting unintentional action in video.
\newblock \emph{CVPR}, 2020.

\bibitem[Farneb{\"a}ck(2003)]{farneback2003two}
Gunnar Farneb{\"a}ck.
\newblock Two-frame motion estimation based on polynomial expansion.
\newblock In \emph{Image Analysis: 13th Scandinavian Conference, SCIA 2003 Halmstad, Sweden, June 29--July 2, 2003 Proceedings 13}, pp.\  363--370. Springer, 2003.

\bibitem[Fu et~al.(2024)Fu, Dai, Luo, Li, Ren, Zhang, Wang, Zhou, Shen, Zhang, Chen, Li, Lin, Zhao, Li, Xu, Zheng, Chen, Ji, and Sun]{fu2024videommefirstevercomprehensiveevaluation}
Chaoyou Fu, Yuhan Dai, Yongdong Luo, Lei Li, Shuhuai Ren, Renrui Zhang, Zihan Wang, Chenyu Zhou, Yunhang Shen, Mengdan Zhang, Peixian Chen, Yanwei Li, Shaohui Lin, Sirui Zhao, Ke~Li, Tong Xu, Xiawu Zheng, Enhong Chen, Rongrong Ji, and Xing Sun.
\newblock Video-mme: The first-ever comprehensive evaluation benchmark of multi-modal llms in video analysis, 2024.
\newblock URL \url{https://arxiv.org/abs/2405.21075}.

\bibitem[Furman \& Wang(2008)Furman and Wang]{furman2008similarity}
Moran Furman and Xiao-Jing Wang.
\newblock Similarity effect and optimal control of multiple-choice decision making.
\newblock \emph{Neuron}, 60\penalty0 (6):\penalty0 1153--1168, 2008.

\bibitem[Gao et~al.(2017)Gao, Sun, Yang, and Nevatia]{gao2017tall}
Jiyang Gao, Chen Sun, Zhenheng Yang, and Ram Nevatia.
\newblock Tall: Temporal activity localization via language query.
\newblock In \emph{Proceedings of the IEEE international conference on computer vision}, pp.\  5267--5275, 2017.

\bibitem[{Gemini Team}(2024)]{geminiteam2024gemini}
{Gemini Team}.
\newblock Gemini: A family of highly capable multimodal models, 2024.

\bibitem[Girdhar et~al.(2023)Girdhar, El-Nouby, Liu, Singh, Alwala, Joulin, and Misra]{girdhar2023imagebind}
Rohit Girdhar, Alaaeldin El-Nouby, Zhuang Liu, Mannat Singh, Kalyan~Vasudev Alwala, Armand Joulin, and Ishan Misra.
\newblock Imagebind: One embedding space to bind them all.
\newblock In \emph{CVPR}, 2023.

\bibitem[Grauman et~al.(2024)Grauman, Westbury, Torresani, Kitani, Malik, Afouras, Ashutosh, Baiyya, Bansal, Boote, et~al.]{grauman2024ego}
Kristen Grauman, Andrew Westbury, Lorenzo Torresani, Kris Kitani, Jitendra Malik, Triantafyllos Afouras, Kumar Ashutosh, Vijay Baiyya, Siddhant Bansal, Bikram Boote, et~al.
\newblock Ego-exo4d: Understanding skilled human activity from first-and third-person perspectives.
\newblock In \emph{Proceedings of the IEEE/CVF Conference on Computer Vision and Pattern Recognition}, pp.\  19383--19400, 2024.

\bibitem[Gurari et~al.(2018)Gurari, Li, Stangl, Guo, Lin, Grauman, Luo, and Bigham]{gurari2018vizwiz}
Danna Gurari, Qing Li, Abigale~J Stangl, Anhong Guo, Chi Lin, Kristen Grauman, Jiebo Luo, and Jeffrey~P Bigham.
\newblock Vizwiz grand challenge: Answering visual questions from blind people.
\newblock In \emph{Proceedings of the IEEE conference on computer vision and pattern recognition}, pp.\  3608--3617, 2018.

\bibitem[He et~al.(2024{\natexlab{a}})He, Li, Jang, Jia, Cao, Shah, Shrivastava, and Lim]{he2024malmm}
Bo~He, Hengduo Li, Young~Kyun Jang, Menglin Jia, Xuefei Cao, Ashish Shah, Abhinav Shrivastava, and Ser-Nam Lim.
\newblock Ma-lmm: Memory-augmented large multimodal model for long-term video understanding.
\newblock In \emph{Proceedings of the IEEE/CVF Conference on Computer Vision and Pattern Recognition (CVPR)}, 2024{\natexlab{a}}.

\bibitem[He et~al.(2024{\natexlab{b}})He, Feng, Zheng, Lu, Zhu, Li, Fan, Wang, Li, Yang, Lin, Wang, Wang, and Wang]{he2024mmworldmultidisciplinemultifacetedworld}
Xuehai He, Weixi Feng, Kaizhi Zheng, Yujie Lu, Wanrong Zhu, Jiachen Li, Yue Fan, Jianfeng Wang, Linjie Li, Zhengyuan Yang, Kevin Lin, William~Yang Wang, Lijuan Wang, and Xin~Eric Wang.
\newblock Mmworld: Towards multi-discipline multi-faceted world model evaluation in videos, 2024{\natexlab{b}}.
\newblock URL \url{https://arxiv.org/abs/2406.08407}.

\bibitem[Hong et~al.(2023)Hong, Zhen, Chen, Zheng, Du, Chen, and Gan]{3dllm}
Yining Hong, Haoyu Zhen, Peihao Chen, Shuhong Zheng, Yilun Du, Zhenfang Chen, and Chuang Gan.
\newblock 3d-llm: Injecting the 3d world into large language models.
\newblock \emph{NeurIPS}, 2023.

\bibitem[Hudson \& Manning(2019)Hudson and Manning]{hudson2019gqa}
Drew~A Hudson and Christopher~D Manning.
\newblock Gqa: A new dataset for real-world visual reasoning and compositional question answering.
\newblock In \emph{Proceedings of the IEEE/CVF conference on computer vision and pattern recognition}, pp.\  6700--6709, 2019.

\bibitem[Jang et~al.(2017)Jang, Song, Yu, Kim, and Kim]{jang2017tgif}
Yunseok Jang, Yale Song, Youngjae Yu, Youngjin Kim, and Gunhee Kim.
\newblock Tgif-qa: Toward spatio-temporal reasoning in visual question answering.
\newblock In \emph{Proceedings of the IEEE conference on computer vision and pattern recognition}, pp.\  2758--2766, 2017.

\bibitem[Kamradt(2023)]{kam2023}
Gregory Kamradt.
\newblock Needle in a haystack - pressure testing llms.
\newblock \url{https://github.com/gkamradt/LLMTest_NeedleInAHaystack}, 2023.
\newblock Accessed: 2024-10-01.

\bibitem[Kim et~al.(2024)Kim, Choi, Lee, and Rhee]{kim2024image}
Wonkyun Kim, Changin Choi, Wonseok Lee, and Wonjong Rhee.
\newblock An image grid can be worth a video: Zero-shot video question answering using a vlm.
\newblock \emph{arXiv preprint arXiv:2403.18406}, 2024.

\bibitem[Krishna et~al.(2017)Krishna, Hata, Ren, Fei-Fei, and Carlos~Niebles]{krishna2017dense}
Ranjay Krishna, Kenji Hata, Frederic Ren, Li~Fei-Fei, and Juan Carlos~Niebles.
\newblock Dense-captioning events in videos.
\newblock In \emph{Proceedings of the IEEE international conference on computer vision}, pp.\  706--715, 2017.

\bibitem[Lei et~al.(2023)Lei, Berg, and Bansal]{lei2022revealing}
Jie Lei, Tamara Berg, and Mohit Bansal.
\newblock Revealing single frame bias for video-and-language learning.
\newblock In Anna Rogers, Jordan Boyd-Graber, and Naoaki Okazaki (eds.), \emph{Proceedings of the 61st Annual Meeting of the Association for Computational Linguistics (Volume 1: Long Papers)}, pp.\  487--507, Toronto, Canada, July 2023. Association for Computational Linguistics.
\newblock \doi{10.18653/v1/2023.acl-long.29}.
\newblock URL \url{https://aclanthology.org/2023.acl-long.29}.

\bibitem[Li et~al.(2024{\natexlab{a}})Li, Zhang, Guo, Zhang, Li, Zhang, Zhang, Li, Liu, and Li]{li2024llavaonevision}
Bo~Li, Yuanhan Zhang, Dong Guo, Renrui Zhang, Feng Li, Hao Zhang, Kaichen Zhang, Yanwei Li, Ziwei Liu, and Chunyuan Li.
\newblock Llava-onevision: Easy visual task transfer.
\newblock \emph{arXiv preprint arXiv:2408.03326}, 2024{\natexlab{a}}.

\bibitem[Li et~al.(2023{\natexlab{a}})Li, Wang, Wang, Ge, Ge, and Shan]{li2023seed}
Bohao Li, Rui Wang, Guangzhi Wang, Yuying Ge, Yixiao Ge, and Ying Shan.
\newblock Seed-bench: Benchmarking multimodal llms with generative comprehension.
\newblock \emph{arXiv preprint arXiv:2307.16125}, 2023{\natexlab{a}}.

\bibitem[Li et~al.(2024{\natexlab{b}})Li, Wang, He, Li, Wang, Liu, Wang, Xu, Chen, Luo, Wang, and Qiao]{li2024mvbenchcomprehensivemultimodalvideo}
Kunchang Li, Yali Wang, Yinan He, Yizhuo Li, Yi~Wang, Yi~Liu, Zun Wang, Jilan Xu, Guo Chen, Ping Luo, Limin Wang, and Yu~Qiao.
\newblock Mvbench: A comprehensive multi-modal video understanding benchmark, 2024{\natexlab{b}}.
\newblock URL \url{https://arxiv.org/abs/2311.17005}.

\bibitem[Li et~al.(2023{\natexlab{b}})Li, Du, Zhou, Wang, Zhao, and Wen]{li2023pope}
Yifan Li, Yifan Du, Kun Zhou, Jinpeng Wang, Xin Zhao, and Ji-Rong Wen.
\newblock Evaluating object hallucination in large vision-language models.
\newblock In Houda Bouamor, Juan Pino, and Kalika Bali (eds.), \emph{Proceedings of the 2023 Conference on Empirical Methods in Natural Language Processing}, pp.\  292--305, Singapore, December 2023{\natexlab{b}}. Association for Computational Linguistics.
\newblock \doi{10.18653/v1/2023.emnlp-main.20}.
\newblock URL \url{https://aclanthology.org/2023.emnlp-main.20}.

\bibitem[Lin et~al.(2023)Lin, Zhu, Ye, Ning, Jin, and Yuan]{lin2023video}
Bin Lin, Bin Zhu, Yang Ye, Munan Ning, Peng Jin, and Li~Yuan.
\newblock Video-llava: Learning united visual representation by alignment before projection.
\newblock \emph{arXiv preprint arXiv:2311.10122}, 2023.

\bibitem[Lin(2004)]{lin-2004-rouge}
Chin-Yew Lin.
\newblock {ROUGE}: A package for automatic evaluation of summaries.
\newblock In \emph{Text Summarization Branches Out}, pp.\  74--81, Barcelona, Spain, July 2004. Association for Computational Linguistics.
\newblock URL \url{https://aclanthology.org/W04-1013}.

\bibitem[Lin et~al.(2014)Lin, Maire, Belongie, Hays, Perona, Ramanan, Doll{\'a}r, and Zitnick]{lin2014microsoft}
Tsung-Yi Lin, Michael Maire, Serge Belongie, James Hays, Pietro Perona, Deva Ramanan, Piotr Doll{\'a}r, and C~Lawrence Zitnick.
\newblock Microsoft coco: Common objects in context.
\newblock In \emph{Computer Vision--ECCV 2014: 13th European Conference, Zurich, Switzerland, September 6-12, 2014, Proceedings, Part V 13}, pp.\  740--755. Springer, 2014.

\bibitem[Liu et~al.(2023{\natexlab{a}})Liu, Li, Wu, and Lee]{liu2023llava}
Haotian Liu, Chunyuan Li, Qingyang Wu, and Yong~Jae Lee.
\newblock Visual instruction tuning.
\newblock \emph{NeurIPS}, 2023{\natexlab{a}}.

\bibitem[Liu et~al.(2024{\natexlab{a}})Liu, Li, Li, and Lee]{liu2023improvedllava}
Haotian Liu, Chunyuan Li, Yuheng Li, and Yong~Jae Lee.
\newblock Improved baselines with visual instruction tuning, 2024{\natexlab{a}}.

\bibitem[Liu et~al.(2024{\natexlab{b}})Liu, Li, Li, Li, Zhang, Shen, and Lee]{liu2024llavanext}
Haotian Liu, Chunyuan Li, Yuheng Li, Bo~Li, Yuanhan Zhang, Sheng Shen, and Yong~Jae Lee.
\newblock Llava-next: Improved reasoning, ocr, and world knowledge, January 2024{\natexlab{b}}.
\newblock URL \url{https://llava-vl.github.io/blog/2024-01-30-llava-next/}.

\bibitem[Liu et~al.(2023{\natexlab{b}})Liu, Duan, Zhang, Li, Zhang, Zhao, Yuan, Wang, He, Liu, et~al.]{liu2023mmbench}
Yuan Liu, Haodong Duan, Yuanhan Zhang, Bo~Li, Songyang Zhang, Wangbo Zhao, Yike Yuan, Jiaqi Wang, Conghui He, Ziwei Liu, et~al.
\newblock Mmbench: Is your multi-modal model an all-around player?
\newblock \emph{arXiv preprint arXiv:2307.06281}, 2023{\natexlab{b}}.

\bibitem[Liu et~al.(2024{\natexlab{c}})Liu, Li, Liu, Wang, Ren, Li, Chen, Sun, and Hou]{liu2024tempcompassvideollmsreally}
Yuanxin Liu, Shicheng Li, Yi~Liu, Yuxiang Wang, Shuhuai Ren, Lei Li, Sishuo Chen, Xu~Sun, and Lu~Hou.
\newblock Tempcompass: Do video llms really understand videos?, 2024{\natexlab{c}}.
\newblock URL \url{https://arxiv.org/abs/2403.00476}.

\bibitem[Lu et~al.(2024)Lu, Bansal, Xia, Liu, Li, Hajishirzi, Cheng, Chang, Galley, and Gao]{lu2024mathvistaevaluatingmathematicalreasoning}
Pan Lu, Hritik Bansal, Tony Xia, Jiacheng Liu, Chunyuan Li, Hannaneh Hajishirzi, Hao Cheng, Kai-Wei Chang, Michel Galley, and Jianfeng Gao.
\newblock Mathvista: Evaluating mathematical reasoning of foundation models in visual contexts, 2024.
\newblock URL \url{https://arxiv.org/abs/2310.02255}.

\bibitem[Mangalam et~al.(2024)Mangalam, Akshulakov, and Malik]{mangalam2023egoschema}
Karttikeya Mangalam, Raiymbek Akshulakov, and Jitendra Malik.
\newblock Egoschema: A diagnostic benchmark for very long-form video language understanding.
\newblock In \emph{Adv. Neural Inform. Process. Syst.}, 2024.

\bibitem[Mathew et~al.(2021)Mathew, Karatzas, and Jawahar]{mathew2021docvqa}
Minesh Mathew, Dimosthenis Karatzas, and CV~Jawahar.
\newblock Docvqa: A dataset for vqa on document images.
\newblock In \emph{Proceedings of the IEEE/CVF winter conference on applications of computer vision}, pp.\  2200--2209, 2021.

\bibitem[Mathew et~al.(2022)Mathew, Bagal, Tito, Karatzas, Valveny, and Jawahar]{mathew2021infographicvqa}
Minesh Mathew, Viraj Bagal, Rub{\`e}n Tito, Dimosthenis Karatzas, Ernest Valveny, and CV~Jawahar.
\newblock Infographicvqa.
\newblock In \emph{Proceedings of the IEEE/CVF Winter Conference on Applications of Computer Vision}, pp.\  1697--1706, 2022.

\bibitem[Meta(2024)]{llama-3}
Meta.
\newblock Llama-3.
\newblock \url{https://ai.meta.com/blog/meta-llama-3/}, 2024.

\bibitem[Ni et~al.(2022)Ni, Peng, Chen, Zhang, Meng, Fu, Xiang, and Ling]{xclip}
Bolin Ni, Houwen Peng, Minghao Chen, Songyang Zhang, Gaofeng Meng, Jianlong Fu, Shiming Xiang, and Haibin Ling.
\newblock Expanding language-image pretrained models for general video recognition.
\newblock In \emph{European Conference on Computer Vision (ECCV)}, 2022.

\bibitem[OpenAI(2023{\natexlab{a}})]{GPT4V_System_Card}
OpenAI.
\newblock Gpt-4v(ision) system card.
\newblock \url{https://cdn.openai.com/papers/GPTV_System_Card.pdf}, 2023{\natexlab{a}}.

\bibitem[OpenAI(2023{\natexlab{b}})]{chatgpt}
OpenAI.
\newblock Chatgpt.
\newblock \url{https://openai.com/blog/chatgpt/}, 2023{\natexlab{b}}.

\bibitem[OpenAI(2023{\natexlab{c}})]{gpt4}
OpenAI.
\newblock Gpt-4 technical report.
\newblock 2023{\natexlab{c}}.

\bibitem[OpenAI(2024)]{GPT4o}
OpenAI.
\newblock Gpt-4o.
\newblock \url{https://openai.com/index/hello-gpt-4o/}, 2024.

\bibitem[Papineni et~al.(2002)Papineni, Roukos, Ward, and Zhu]{papineni-etal-2002-bleu}
Kishore Papineni, Salim Roukos, Todd Ward, and Wei-Jing Zhu.
\newblock {B}leu: a method for automatic evaluation of machine translation.
\newblock In Pierre Isabelle, Eugene Charniak, and Dekang Lin (eds.), \emph{Proceedings of the 40th Annual Meeting of the Association for Computational Linguistics}, pp.\  311--318, Philadelphia, Pennsylvania, USA, July 2002. Association for Computational Linguistics.
\newblock \doi{10.3115/1073083.1073135}.
\newblock URL \url{https://aclanthology.org/P02-1040}.

\bibitem[Peng et~al.(2023)Peng, Wang, Dong, Hao, Huang, Ma, and Wei]{peng2023kosmos}
Zhiliang Peng, Wenhui Wang, Li~Dong, Yaru Hao, Shaohan Huang, Shuming Ma, and Furu Wei.
\newblock Kosmos-2: Grounding multimodal large language models to the world.
\newblock \emph{arXiv preprint arXiv:2306.14824}, 2023.

\bibitem[Radford et~al.(2021)Radford, Kim, Hallacy, Ramesh, Goh, Agarwal, Sastry, Askell, Mishkin, Clark, et~al.]{radford2021learning}
Alec Radford, Jong~Wook Kim, Chris Hallacy, Aditya Ramesh, Gabriel Goh, Sandhini Agarwal, Girish Sastry, Amanda Askell, Pamela Mishkin, Jack Clark, et~al.
\newblock Learning transferable visual models from natural language supervision.
\newblock In \emph{International conference on machine learning}, pp.\  8748--8763. PMLR, 2021.

\bibitem[Reimers \& Gurevych(2019)Reimers and Gurevych]{reimers-2019-sentence-bert}
Nils Reimers and Iryna Gurevych.
\newblock Sentence-bert: Sentence embeddings using siamese bert-networks.
\newblock In \emph{Proceedings of the 2019 Conference on Empirical Methods in Natural Language Processing}. Association for Computational Linguistics, 11 2019.
\newblock URL \url{https://arxiv.org/abs/1908.10084}.

\bibitem[Rohrbach et~al.(2015)Rohrbach, Rohrbach, Tandon, and Schiele]{rohrbach15cvpr}
Anna Rohrbach, Marcus Rohrbach, Niket Tandon, and Bernt Schiele.
\newblock A dataset for movie description.
\newblock In \emph{Proceedings of the IEEE Conference on Computer Vision and Pattern Recognition (CVPR)}, 2015.

\bibitem[Shao et~al.(2020)Shao, Zhao, Dai, and Lin]{shao2020finegym}
Dian Shao, Yue Zhao, Bo~Dai, and Dahua Lin.
\newblock Finegym: A hierarchical video dataset for fine-grained action understanding.
\newblock In \emph{IEEE Conference on Computer Vision and Pattern Recognition (CVPR)}, 2020.

\bibitem[Singh et~al.(2019)Singh, Natarajan, Shah, Jiang, Chen, Batra, Parikh, and Rohrbach]{singh2019textvqa}
Amanpreet Singh, Vivek Natarajan, Meet Shah, Yu~Jiang, Xinlei Chen, Dhruv Batra, Devi Parikh, and Marcus Rohrbach.
\newblock Towards vqa models that can read.
\newblock In \emph{Proceedings of the IEEE/CVF conference on computer vision and pattern recognition}, pp.\  8317--8326, 2019.

\bibitem[Sun et~al.(2023)Sun, Shen, Cao, Liu, Li, Shen, Gan, Gui, Wang, Yang, et~al.]{sun2023aligning}
Zhiqing Sun, Sheng Shen, Shengcao Cao, Haotian Liu, Chunyuan Li, Yikang Shen, Chuang Gan, Liang-Yan Gui, Yu-Xiong Wang, Yiming Yang, et~al.
\newblock Aligning large multimodal models with factually augmented rlhf.
\newblock \emph{arXiv preprint arXiv:2309.14525}, 2023.

\bibitem[Tan et~al.(2024)Tan, Sun, Hu, Wang, Deilamsalehy, Plummer, Russell, and Saenko]{tan2024koala}
Reuben Tan, Ximeng Sun, Ping Hu, Jui-hsien Wang, Hanieh Deilamsalehy, Bryan~A Plummer, Bryan Russell, and Kate Saenko.
\newblock Koala: Key frame-conditioned long video-llm.
\newblock In \emph{Proceedings of the IEEE/CVF Conference on Computer Vision and Pattern Recognition}, pp.\  13581--13591, 2024.

\bibitem[Tang et~al.(2019)Tang, Ding, Rao, Zheng, Zhang, Zhao, Lu, and Zhou]{tang2019coin}
Yansong Tang, Dajun Ding, Yongming Rao, Yu~Zheng, Danyang Zhang, Lili Zhao, Jiwen Lu, and Jie Zhou.
\newblock Coin: A large-scale dataset for comprehensive instructional video analysis.
\newblock In \emph{Proceedings of the IEEE/CVF Conference on Computer Vision and Pattern Recognition}, pp.\  1207--1216, 2019.

\bibitem[Touvron et~al.(2023)Touvron, Lavril, Izacard, Martinet, Lachaux, Lacroix, Rozi{\`e}re, Goyal, Hambro, Azhar, et~al.]{touvron2023llama}
Hugo Touvron, Thibaut Lavril, Gautier Izacard, Xavier Martinet, Marie-Anne Lachaux, Timoth{\'e}e Lacroix, Baptiste Rozi{\`e}re, Naman Goyal, Eric Hambro, Faisal Azhar, et~al.
\newblock Llama: Open and efficient foundation language models.
\newblock \emph{arXiv preprint arXiv:2302.13971}, 2023.

\bibitem[Vedantam et~al.(2015)Vedantam, Lawrence~Zitnick, and Parikh]{vedantam2015cider}
Ramakrishna Vedantam, C~Lawrence~Zitnick, and Devi Parikh.
\newblock Cider: Consensus-based image description evaluation.
\newblock In \emph{Proceedings of the IEEE conference on computer vision and pattern recognition}, pp.\  4566--4575, 2015.

\bibitem[Wang et~al.(2024)Wang, Bai, Tan, Wang, Fan, Bai, Chen, Liu, Wang, Ge, Fan, Dang, Du, Ren, Men, Liu, Zhou, Zhou, and Lin]{Qwen2VL}
Peng Wang, Shuai Bai, Sinan Tan, Shijie Wang, Zhihao Fan, Jinze Bai, Keqin Chen, Xuejing Liu, Jialin Wang, Wenbin Ge, Yang Fan, Kai Dang, Mengfei Du, Xuancheng Ren, Rui Men, Dayiheng Liu, Chang Zhou, Jingren Zhou, and Junyang Lin.
\newblock Qwen2-vl: Enhancing vision-language model's perception of the world at any resolution.
\newblock \emph{arXiv preprint arXiv:2409.12191}, 2024.

\bibitem[Wei et~al.(2021)Wei, Bosma, Zhao, Guu, Yu, Lester, Du, Dai, and Le]{weifinetuned}
Jason Wei, Maarten Bosma, Vincent Zhao, Kelvin Guu, Adams~Wei Yu, Brian Lester, Nan Du, Andrew~M Dai, and Quoc~V Le.
\newblock Finetuned language models are zero-shot learners.
\newblock In \emph{International Conference on Learning Representations}, 2021.

\bibitem[Wu(2024)]{wu2024freeva}
Wenhao Wu.
\newblock Freeva: Offline mllm as training-free video assistant.
\newblock \emph{arXiv preprint arXiv:2405.07798}, 2024.

\bibitem[Xiao et~al.(2021)Xiao, Shang, Yao, and Chua]{xiao2021next}
Junbin Xiao, Xindi Shang, Angela Yao, and Tat-Seng Chua.
\newblock Next-qa: Next phase of question-answering to explaining temporal actions.
\newblock In \emph{Proceedings of the IEEE/CVF conference on computer vision and pattern recognition}, pp.\  9777--9786, 2021.

\bibitem[Xu et~al.(2017)Xu, Zhao, Xiao, Wu, Zhang, He, and Zhuang]{xu2017video}
Dejing Xu, Zhou Zhao, Jun Xiao, Fei Wu, Hanwang Zhang, Xiangnan He, and Yueting Zhuang.
\newblock Video question answering via gradually refined attention over appearance and motion.
\newblock In \emph{Proceedings of the 25th ACM international conference on Multimedia}, pp.\  1645--1653, 2017.

\bibitem[Xu et~al.(2021)Xu, Ghosh, Huang, Okhonko, Aghajanyan, Metze, Zettlemoyer, and Feichtenhofer]{xu-etal-2021-videoclip}
Hu~Xu, Gargi Ghosh, Po-Yao Huang, Dmytro Okhonko, Armen Aghajanyan, Florian Metze, Luke Zettlemoyer, and Christoph Feichtenhofer.
\newblock {V}ideo{CLIP}: Contrastive pre-training for zero-shot video-text understanding.
\newblock In \emph{EMNLP}, 2021.

\bibitem[Xu et~al.(2016)Xu, Mei, Yao, and Rui]{xu2016msr}
Jun Xu, Tao Mei, Ting Yao, and Yong Rui.
\newblock Msr-vtt: A large video description dataset for bridging video and language.
\newblock In \emph{Proceedings of the IEEE conference on computer vision and pattern recognition}, pp.\  5288--5296, 2016.

\bibitem[Yao et~al.(2024)Yao, Yu, Zhang, Wang, Cui, Zhu, Cai, Li, Zhao, He, et~al.]{yao2024minicpm}
Yuan Yao, Tianyu Yu, Ao~Zhang, Chongyi Wang, Junbo Cui, Hongji Zhu, Tianchi Cai, Haoyu Li, Weilin Zhao, Zhihui He, et~al.
\newblock Minicpm-v: A gpt-4v level mllm on your phone.
\newblock \emph{arXiv preprint arXiv:2408.01800}, 2024.

\bibitem[Young et~al.(2024)Young, Chen, Li, Huang, Zhang, Zhang, Li, Zhu, Chen, Chang, Yu, Liu, Liu, Yue, Yang, Yang, Yu, Xie, Huang, Hu, Ren, Niu, Nie, Xu, Liu, Wang, Cai, Gu, Liu, and Dai]{yi}
Alex Young, Bei Chen, Chao Li, Chengen Huang, Ge~Zhang, Guanwei Zhang, Heng Li, Jiangcheng Zhu, Jianqun Chen, Jing Chang, Kaidong Yu, Peng Liu, Qiang Liu, Shawn Yue, Senbin Yang, Shiming Yang, Tao Yu, Wen Xie, Wenhao Huang, Xiaohui Hu, Xiaoyi Ren, Xinyao Niu, Pengcheng Nie, Yuchi Xu, Yudong Liu, Yue Wang, Yuxuan Cai, Zhenyu Gu, Zhiyuan Liu, and Zonghong Dai.
\newblock Yi: Open foundation models by 01.ai.
\newblock \emph{arXiv}, 2024.

\bibitem[Young et~al.(2014)Young, Lai, Hodosh, and Hockenmaier]{young2014image}
Peter Young, Alice Lai, Micah Hodosh, and Julia Hockenmaier.
\newblock From image descriptions to visual denotations: New similarity metrics for semantic inference over event descriptions.
\newblock \emph{Transactions of the Association for Computational Linguistics}, 2:\penalty0 67--78, 2014.

\bibitem[Yu et~al.(2024{\natexlab{a}})Yu, Yao, Zhang, He, Han, Cui, Hu, Liu, Zheng, Sun, et~al.]{yu2024rlhf}
Tianyu Yu, Yuan Yao, Haoye Zhang, Taiwen He, Yifeng Han, Ganqu Cui, Jinyi Hu, Zhiyuan Liu, Hai-Tao Zheng, Maosong Sun, et~al.
\newblock Rlhf-v: Towards trustworthy mllms via behavior alignment from fine-grained correctional human feedback.
\newblock In \emph{Proceedings of the IEEE/CVF Conference on Computer Vision and Pattern Recognition}, pp.\  13807--13816, 2024{\natexlab{a}}.

\bibitem[Yu et~al.(2024{\natexlab{b}})Yu, Yang, Li, Wang, Lin, Liu, Wang, and Wang]{yu2023mmvet}
Weihao Yu, Zhengyuan Yang, Linjie Li, Jianfeng Wang, Kevin Lin, Zicheng Liu, Xinchao Wang, and Lijuan Wang.
\newblock Mm-vet: Evaluating large multimodal models for integrated capabilities.
\newblock In \emph{Forty-first International Conference on Machine Learning}, 2024{\natexlab{b}}.

\bibitem[Yu et~al.(2019{\natexlab{a}})Yu, Xu, Yu, Yu, Zhao, Zhuang, and Tao]{yu2019activitynet}
Zhou Yu, Dejing Xu, Jun Yu, Ting Yu, Zhou Zhao, Yueting Zhuang, and Dacheng Tao.
\newblock Activitynet-qa: A dataset for understanding complex web videos via question answering.
\newblock In \emph{AAAI}, volume~33, pp.\  9127--9134, 2019{\natexlab{a}}.

\bibitem[Yu et~al.(2019{\natexlab{b}})Yu, Xu, Yu, Yu, Zhao, Zhuang, and Tao]{yu2019activitynetqadatasetunderstandingcomplex}
Zhou Yu, Dejing Xu, Jun Yu, Ting Yu, Zhou Zhao, Yueting Zhuang, and Dacheng Tao.
\newblock Activitynet-qa: A dataset for understanding complex web videos via question answering, 2019{\natexlab{b}}.
\newblock URL \url{https://arxiv.org/abs/1906.02467}.

\bibitem[Yue et~al.(2024{\natexlab{a}})Yue, Ni, Zhang, Zheng, Liu, Zhang, Stevens, Jiang, Ren, Sun, Wei, Yu, Yuan, Sun, Yin, Zheng, Yang, Liu, Huang, Sun, Su, and Chen]{yue2023mmmu}
Xiang Yue, Yuansheng Ni, Kai Zhang, Tianyu Zheng, Ruoqi Liu, Ge~Zhang, Samuel Stevens, Dongfu Jiang, Weiming Ren, Yuxuan Sun, Cong Wei, Botao Yu, Ruibin Yuan, Renliang Sun, Ming Yin, Boyuan Zheng, Zhenzhu Yang, Yibo Liu, Wenhao Huang, Huan Sun, Yu~Su, and Wenhu Chen.
\newblock Mmmu: A massive multi-discipline multimodal understanding and reasoning benchmark for expert agi.
\newblock In \emph{Proceedings of CVPR}, 2024{\natexlab{a}}.

\bibitem[Yue et~al.(2024{\natexlab{b}})Yue, Zheng, Ni, Wang, Zhang, Tong, Sun, Yin, Yu, Zhang, et~al.]{yue2024mmmupro}
Xiang Yue, Tianyu Zheng, Yuansheng Ni, Yubo Wang, Kai Zhang, Shengbang Tong, Yuxuan Sun, Ming Yin, Botao Yu, Ge~Zhang, et~al.
\newblock Mmmu-pro: A more robust multi-discipline multimodal understanding benchmark.
\newblock \emph{arXiv preprint arXiv:2409.02813}, 2024{\natexlab{b}}.

\bibitem[Zhang et~al.(2019)Zhang, Dai, and Wang]{zhang2019dynamictemporalpyramidnetwork}
Da~Zhang, Xiyang Dai, and Yuan-Fang Wang.
\newblock Dynamic temporal pyramid network: A closer look at multi-scale modeling for activity detection.
\newblock In \emph{Computer Vision--ACCV 2018: 14th Asian Conference on Computer Vision, Perth, Australia, December 2--6, 2018, Revised Selected Papers, Part IV 14}, pp.\  712--728. Springer, 2019.

\bibitem[Zhang et~al.(2023{\natexlab{a}})Zhang, Li, and Bing]{zhang2023video}
Hang Zhang, Xin Li, and Lidong Bing.
\newblock Video-llama: An instruction-tuned audio-visual language model for video understanding.
\newblock \emph{arXiv preprint arXiv:2306.02858}, 2023{\natexlab{a}}.

\bibitem[Zhang et~al.(2023{\natexlab{b}})Zhang, Li, Li, Ren, Zou, Liu, Huang, Gao, Zhang, Li, and Yang]{zhang2023llavagrounding}
Hao Zhang, Hongyang Li, Feng Li, Tianhe Ren, Xueyan Zou, Shilong Liu, Shijia Huang, Jianfeng Gao, Lei Zhang, Chunyuan Li, and Jianwei Yang.
\newblock Llava-grounding: Grounded visual chat with large multimodal models, 2023{\natexlab{b}}.

\bibitem[Zhang et~al.(2024{\natexlab{a}})Zhang, Dong, Zang, Cao, Qian, Chen, Guo, Duan, Wang, Ouyang, Zhang, Zhang, Li, Gao, Sun, Zhang, Li, Li, Wang, Yan, He, Zhang, Chen, Dai, Qiao, Lin, and Wang]{internlmxcomposer2_5}
Pan Zhang, Xiaoyi Dong, Yuhang Zang, Yuhang Cao, Rui Qian, Lin Chen, Qipeng Guo, Haodong Duan, Bin Wang, Linke Ouyang, Songyang Zhang, Wenwei Zhang, Yining Li, Yang Gao, Peng Sun, Xinyue Zhang, Wei Li, Jingwen Li, Wenhai Wang, Hang Yan, Conghui He, Xingcheng Zhang, Kai Chen, Jifeng Dai, Yu~Qiao, Dahua Lin, and Jiaqi Wang.
\newblock Internlm-xcomposer-2.5: A versatile large vision language model supporting long-contextual input and output.
\newblock \emph{arXiv preprint arXiv:2407.03320}, 2024{\natexlab{a}}.

\bibitem[Zhang et~al.(2023{\natexlab{c}})Zhang, Sun, Chen, Xiao, Shao, Zhang, Chen, and Luo]{zhang2023gpt4roi}
Shilong Zhang, Peize Sun, Shoufa Chen, Min Xiao, Wenqi Shao, Wenwei Zhang, Kai Chen, and Ping Luo.
\newblock Gpt4roi: Instruction tuning large language model on region-of-interest.
\newblock \emph{arXiv preprint arXiv:2307.03601}, 2023{\natexlab{c}}.

\bibitem[Zhang et~al.(2024{\natexlab{b}})Zhang, Li, Liu, Lee, Gui, Fu, Feng, Liu, and Li]{zhang2024llavanextvideo}
Yuanhan Zhang, Bo~Li, haotian Liu, Yong~jae Lee, Liangke Gui, Di~Fu, Jiashi Feng, Ziwei Liu, and Chunyuan Li.
\newblock Llava-next: A strong zero-shot video understanding model, April 2024{\natexlab{b}}.
\newblock URL \url{https://llava-vl.github.io/blog/2024-04-30-llava-next-video/}.

\bibitem[Zhu et~al.(2024{\natexlab{a}})Zhu, Lin, Ning, Yan, Cui, HongFa, Pang, Jiang, Zhang, Li, et~al.]{zhu2023languagebind}
Bin Zhu, Bin Lin, Munan Ning, Yang Yan, Jiaxi Cui, WANG HongFa, Yatian Pang, Wenhao Jiang, Junwu Zhang, Zongwei Li, et~al.
\newblock Languagebind: Extending video-language pretraining to n-modality by language-based semantic alignment.
\newblock In \emph{The Twelfth International Conference on Learning Representations}, 2024{\natexlab{a}}.

\bibitem[Zhu et~al.(2024{\natexlab{b}})Zhu, Chen, Shen, Li, and Elhoseiny]{zhu2023minigpt}
Deyao Zhu, Jun Chen, Xiaoqian Shen, Xiang Li, and Mohamed Elhoseiny.
\newblock Minigpt-4: Enhancing vision-language understanding with advanced large language models.
\newblock \emph{ICLR}, 2024{\natexlab{b}}.

\end{thebibliography}
\bibliographystyle{iclr2025_conference}

\appendix

\clearpage

\section{Broader Impact}
\label{sec:boarder_impact}
\shortname{}, a comprehensive benchmark for video understanding, has the potential to significantly advance research in this field by offering improved metrics for model evaluation. Our work aims to enhance the temporal reasoning capabilities of future video understanding models. However, the broader impact of more advanced video understanding technologies raises important societal concerns, including the risk of mass surveillance, privacy violations, and the development of harmful applications like autonomous weapons. Therefore, we strongly encourage thoughtful consideration when deploying these models in real-world scenarios to mitigate negative or unintended consequences.

\section{More Visualizations of Our Benchmark}
\begin{figure}[h]
    \centering
    \includegraphics[width=0.8\linewidth]{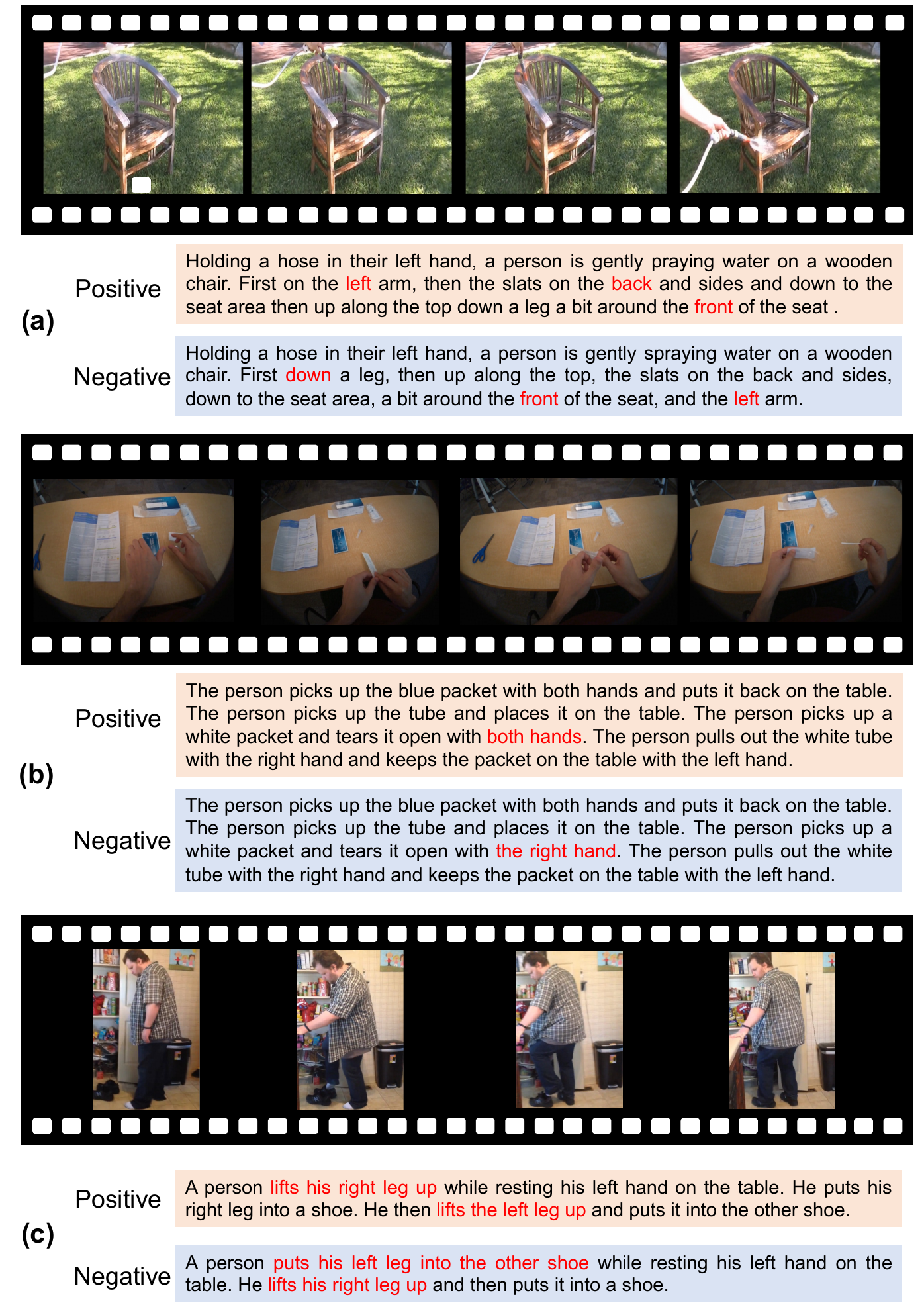}
    \caption{Visualizations (I) of our fine-grained annotations of the videos with both positive and negative descriptions.}
    \vspace{-0.1em}
    \label{fig:example1}
\end{figure}

\begin{figure}[h]
    \centering
    \includegraphics[width=0.8\linewidth]{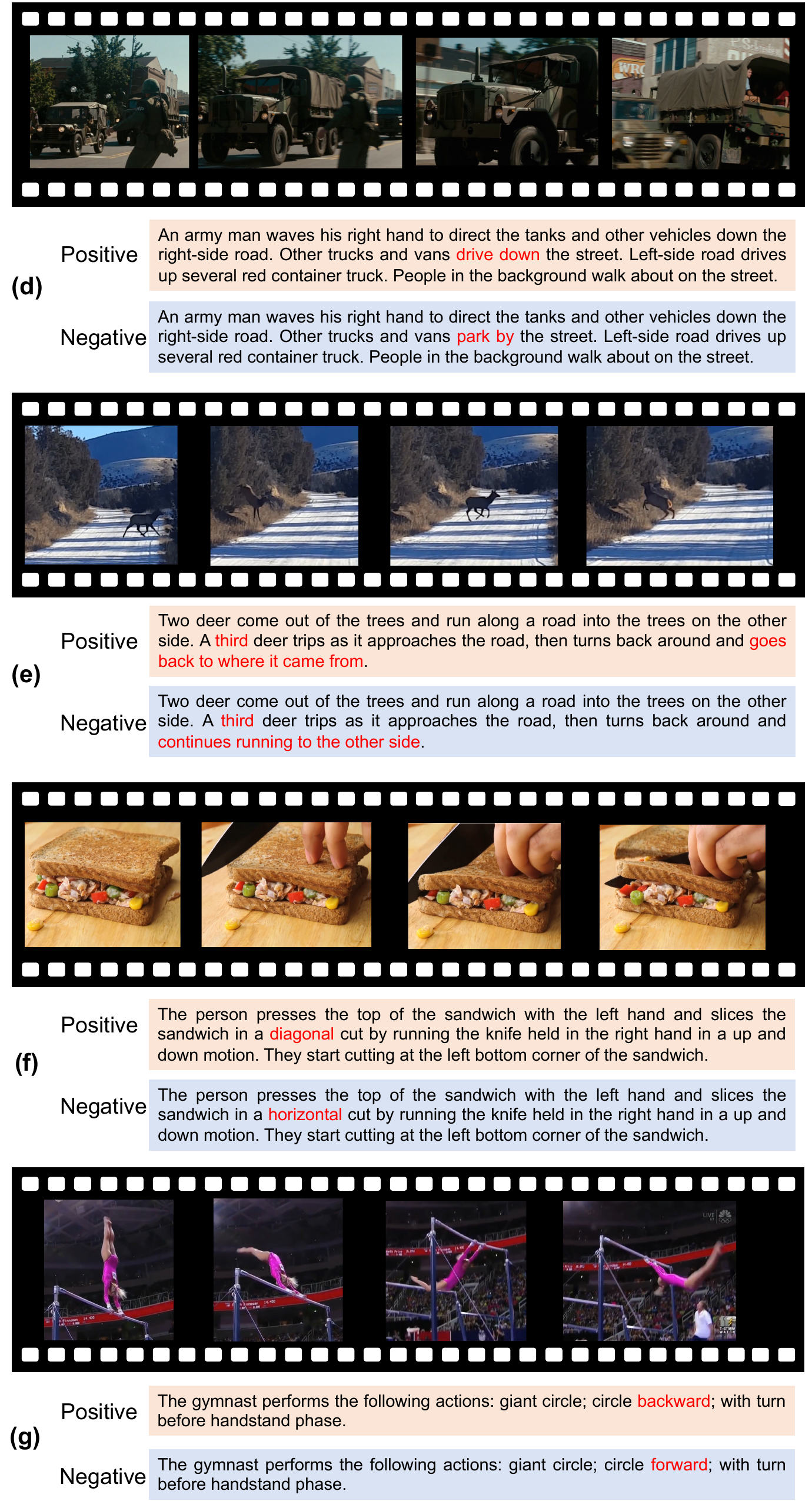}
    \caption{Visualizations (II) of our fine-grained annotations of the videos with both positive and negative descriptions.}
    \vspace{-0.1em}
    \label{fig:example2}
\end{figure}

In this section, we present comprehensive visualizations of our fine-grained annotations with both positive and negative descriptions. For each benchmark mentioned in Table~\ref{tab:data_statistics}, we provide one video example with its positive annotation and one of the corresponding negative descriptions (there are more than one negative for a single video in our dataset) in Figures~\ref{fig:example1}~\&~\ref{fig:example2}. The video examples  (\emph{a - f}) are displayed in the same order as their sources in Table~\ref{tab:data_statistics} (7 in total).

\section{Per subset Results for Short and Long  Video QA  under Binary Accuracy (BA)}

The per subset results (denoted as ``T-") for short and long video QA under Binary Accuracy (BA) are shown in Table~\ref{table:main_results_shortqa_BA}, and Table~\ref{table:longvideo_BA}, respectively. Still, human achieve much better performance than all multimodal videos. Interestingly, both human and  Finegym, the professional subset,

\section{More Results with Extended Frames}
\label{appendix: more results.}

In the main paper, we only report the performance of each multimodal video models with the the number of frams that leads to the best performance. Here we extend the results to show the results of more frames in Table~\ref{tab:short_qa_full_table}.

\begin{table}[ht]
    \centering
    \vspace{-1.5em}
    \caption{\shortname{} performance of various models under binary QA accuracy (BA) and multiple binary QA accuracy (MBA) setting for short and long question answering with different number of frames. ``Overall" denotes the average performance of short and long video QA performance. }
    \label{tab:short_qa_full_table}
    \vspace{-1em}
    \resizebox{0.95\textwidth}{!}{
    \begin{tabular}{lcccccccc}
        \toprule
Model                & \# Frames & Overall MBA & Overall  BA & Short  MBA & Short  BA & Long  MBA & Long  BA & Captioning  \\ \midrule  \midrule
Human Performance    & -                         & -              & -             & 67.9         & 89.7        & -           & -          & -                   \\ \midrule\midrule
Random Chance        & -                         & 9.5            & 50.0          & 9.5          & 50.0        & 9.5         & 50.0       & -                   \\ \midrule  \midrule
XCLIP                & 8                         & 12.0           & 51.7          & 12.9         & 51.6        & 11.1        & 51.7       & -                   \\ \midrule
ImageBind            & 2                         & 12.4           & 52.0          & 14.0         & 53.0        & 10.7        & 51.0       & -                   \\\midrule
LanguageBind         & 8                         & 13.3           & 52.2          & 14.5         & 52.8        & 12.0        & 51.6       & -                   \\\midrule\midrule
GPT-4o               & 64                        & 35.4           & 73.3          & 38.0         & 76.0        & 32.7        & 70.5       & 63.5                \\
                     & 32                        & 32.9           & 71.5          & 38.3         & 75.9        & 27.4        & 67.0       & 63.2                \\
                     & 16                        & 34.3           & 72.8          & 38.5         & 75.7        & 30.1        & 69.8       & 61.3                \\
                     & 8                         & 32.9           & 72.0          & 37.4         & 75.1        & 28.3        & 68.8       & 60.3                \\
                     & 4                         & 31.9           & 71.2          & 35.8         & 74.4        & 28.0        & 68.0       & 58.8                \\
                     & 2                         & 30.3           & 70.3          & 33.3         & 72.7        & 27.3        & 67.8       & 55.3                \\
                     & 1                         & 26.5           & 67.4          & 28.4         & 70.0        & 24.5        & 64.7       & 52.3                \\
                     & 0                         & 27.4           & 67.7          & 26.5         & 67.7        & 28.2        & 67.6       & -                   \\\midrule
Gemini-1.5-Pro       & 1FPS                      & 25.7           & 66.4          & 26.6         & 67.5        & 24.7        & 65.2       & 56.5                \\
                     & 0                         & 18.7           & 60.2          & 16.1         & 58.1        & 21.2        & 62.2       & -                   \\\midrule
Claude-3.5-Sonnet    & 16                        & 23.2           & 64.2          & 23.5         & 65.9        & 22.9        & 62.4       & 54.1                \\
                     & 8                         & 24.1           & 65.1          & 23.6         & 65.5        & 24.5        & 64.6       & 53.1                \\
                     & 4                         & 23.2           & 64.2          & 23.1         & 64.8        & 23.3        & 63.6       & 51.9                \\
                     & 2                         & 21.1           & 62.2          & 21.2         & 61.9        & 20.9        & 62.4       & 48.2                \\
                     & 1                         & 18.7           & 58.9          & 18.4         & 58.5        & 18.9        & 59.3       & 41.0                \\\midrule
Qwen2-VL-72B         & 32                        & 31.7           & 70.2          & 38.3         & 75.8        & 25.0        & 64.5       & 56.1                \\
                     & 16                        & 31.5           & 70.1          & 36.9         & 74.6        & 26.1        & 65.5       & 54.1                \\
                     & 8                         & 30.1           & 68.9          & 34.0         & 73.1        & 26.2        & 64.7       & 51.4                \\
                     & 4                         & 28.6           & 68.4          & 31.2         & 71.5        & 26.0        & 65.3       & 48.3                \\
                     & 2                         & 27.3           & 67.6          & 27.5         & 69.2        & 27.1        & 66.0       & 43.9                \\\midrule
Qwen2-VL-7B          & 32                        & 21.8           & 62.1          & 24.7         & 64.4        & 18.8        & 59.7       & 51.9                \\
                     & 16                        & 21.2           & 61.5          & 23.6         & 63.3        & 18.7        & 59.7       & 50.3                \\
                     & 8                         & 19.2           & 59.4          & 21.1         & 61.1        & 17.2        & 57.7       & 48.4                \\
                     & 4                         & 17.4           & 58.5          & 19.3         & 59.5        & 15.4        & 57.5       & 46.1                \\
                     & 2                         & 16.4           & 56.9          & 17.7         & 57.8        & 15.0        & 56.0       & 42.0                \\\midrule
LLaVA-OneVision-72B  & 32                        & 26.6           & 66.6          & 30.7         & 70.5        & 22.4        & 62.7       & 53.9                \\
                     & 16                        & 27.2           & 67.3          & 32.1         & 71.2        & 22.3        & 63.4       & 54.2                \\
                     & 8                         & 28.1           & 67.9          & 33.0         & 72.1        & 23.1        & 63.6       & 55.0                \\
                     & 4                         & 27.6           & 67.3          & 31.4         & 71.2        & 23.8        & 63.4       & 54.2                \\
                     & 2                         & 25.7           & 66.3          & 29.2         & 69.6        & 22.1        & 63.0       & 51.1                \\
                     & 1                         & 23.3           & 64.1          & 27.1         & 67.9        & 19.5        & 60.2       & 48.6                \\\midrule
LLaVA-OneVision-7B   & 32                        & 18.7           & 59.4          & 21.2         & 61.9        & 16.2        & 56.9       & 50.1                \\
                     & 16                        & 17.9           & 58.8          & 20.1         & 60.9        & 15.6        & 56.6       & 50.4                \\
                     & 8                         & 17.3           & 57.8          & 19.5         & 59.9        & 15.0        & 55.7       & 50.2                \\
                     & 4                         & 16.4           & 56.3          & 18.9         & 58.9        & 13.9        & 53.7       & 49.7                \\
                     & 2                         & 14.8           & 54.4          & 16.8         & 56.1        & 12.7        & 52.7       & 47.1                \\
                     & 1                         & 12.0           & 51.4          & 13.4         & 53.3        & 10.6        & 49.5       & 44.1                \\\midrule
LLaVA-NeXT-Video-34B & 32                        & 19.9           & 61.1          & 22.0         & 64.0        & 17.7        & 58.2       & 53.1                \\
                     & 16                        & 20.3           & 61.1          & 21.8         & 63.7        & 18.7        & 58.4       & 53.3                \\
                     & 8                         & 20.5           & 61.9          & 21.4         & 63.3        & 19.5        & 60.4       & 53.4                \\
                     & 4                         & 20.4           & 61.7          & 20.7         & 63.0        & 20.0        & 60.3       & 52.5                \\
                     & 2                         & 19.9           & 61.2          & 20.0         & 61.8        & 19.7        & 60.6       & 48.9                \\
                     & 1                         & 19.0           & 59.8          & 19.0         & 60.5        & 18.9        & 59.1       & 46.2                \\\midrule
LLaVA-NeXT-Video-7B  & 32                        & 15.9           & 57.1          & 17.3         & 59.5        & 14.5        & 54.7       & 51.6                \\
                     & 16                        & 19.3           & 59.9          & 22.4         & 64.0        & 16.1        & 55.7       & 49.9                \\
                     & 8                         & 20.5           & 61.2          & 23.6         & 65.1        & 17.3        & 57.2       & 50.1                \\
                     & 4                         & 20.0           & 60.7          & 23.0         & 64.2        & 17.0        & 57.2       & 49.2                \\
                     & 2                         & 19.2           & 60.3          & 21.5         & 63.1        & 16.8        & 57.4       & 46.8                \\
                     & 1                         & 17.6           & 59.1          & 19.1         & 62.0        & 16.1        & 56.1       & 44.0                \\\midrule
InternLM-XC2.5       & 1FPS                      & 16.8           & 57.3          & 17.9         & 58.8        & 15.6        & 55.8       & 52.4                \\\midrule
VideoLLaVA           & 8                         & 20.3           & 61.6          & 25.5         & 67.1        & 15.1        & 56.0       & 46.0                \\\midrule
MiniCPM-V2.6         & 1FPS                      & 20.4           & 61.3          & 21.4         & 62.3        & 19.3        & 60.3       & 47.2                \\\midrule
Phi-3.5-Vision       & 32                        & 14.1           & 54.3          & 15.6         & 56.8        & 12.6        & 51.7       & 48.4                \\
                     & 16                        & 14.7           & 55.1          & 15.9         & 57.2        & 13.5        & 53.0       & 48.9                \\
                     & 8                         & 14.9           & 55.6          & 15.9         & 57.4        & 13.8        & 53.7       & 48.3                \\
                     & 4                         & 15.0           & 56.0          & 15.5         & 57.5        & 14.5        & 54.5       & 44.0                \\
                     & 2                         & 15.5           & 56.2          & 16.9         & 58.0        & 14.1        & 54.4       & 42.9                \\
                     & 1                         & 14.5           & 55.9          & 16.5         & 57.7        & 12.5        & 54.0       & 42.1                \\\midrule
MA-LMM               & 4                         & 9.1            & 47.4          & 9.2          & 48.0        & 9.0         & 46.9       & 38.7                \\\midrule
M3                   & 6                         & 13.3           & 54.7          & 14.8         & 56.4        & 11.8        & 53.1       & 47.8                \\\midrule\midrule
LLaVA-1.5-13B        & 1                         & 13.7           & 55.1          & 13.1         & 55.7        & 14.2        & 54.5       & 47.9                \\\midrule
LLaVA-1.5-7B         & 1                         & 15.3           & 56.8          & 18.3         & 60.5        & 12.3        & 53.2       & 45.7                \\\midrule
LLaVA-NeXT-34B       & 1                         & 19.0           & 60.5          & 18.0         & 60.5        & 19.9        & 60.5       & 49.1                \\\midrule
Phi-3-Vision         & 1                         & 15.4           & 55.2          & 15.1         & 54.4        & 15.6        & 56.0       & 42.0                \\\midrule\midrule
Gemini-1.5-Pro       & 0                         & 18.6           & 60.1          & 16.1         & 58.1        & 21.2        & 62.2       & -                   \\\midrule
Yi-34B               & 0                         & 18.5           & 59.7          & 18.7         & 59.9        & 18.4        & 59.5       & -                   \\\midrule
Vicuna7b-1-5         & 0                         & 10.1           & 50.8          & 10.4         & 50.5        & 9.9         & 51.1       & -                   \\\midrule
Flan-T5-XL           & 0                         & 18.6           & 59.0          & 17.9         & 57.9        & 19.4        & 60.1       & -                   \\\midrule
Flan-T5-XXL          & 0                         & 15.9           & 56.0          & 15.1         & 55.1        & 16.7        & 56.9       & -                  
\\ 
        \bottomrule
    \end{tabular}
    }
\end{table}

\section{Data Annotation Platform}

\begin{figure}[h]
    \centering
    \includegraphics[width=\linewidth]{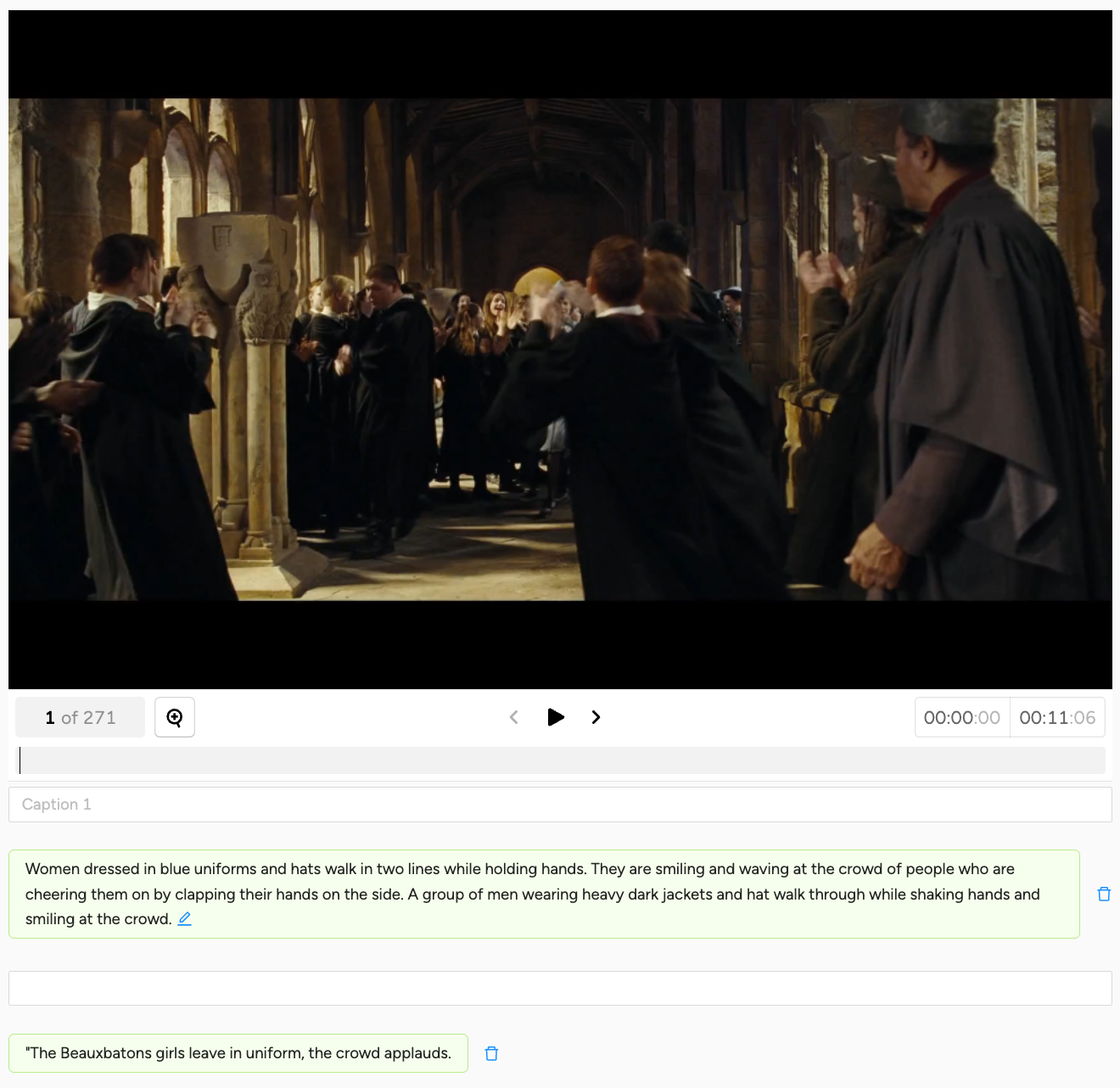}
    \caption{Positive caption refinement platform.}
    \vspace{-0.1em}
    \label{fig:example_pos_cap_refine}
\end{figure}

\paragraph{Positive Captions}
We use Amazon Mechanical Turk (AMT)~\footnote{https://www.mturk.com/} for positive caption annotation, and then use Label Studio~\footnote{https://labelstud.io/} to let authors refine the caption. As shown in Figure~\ref{fig:example_pos_cap_refine}, authors can edit the caption from AMT workers. Also, we provide the original short video  captions to let people better understand our task.

\paragraph{Negative Captions}

\begin{figure}[h]
    \centering
    \includegraphics[width=\linewidth]{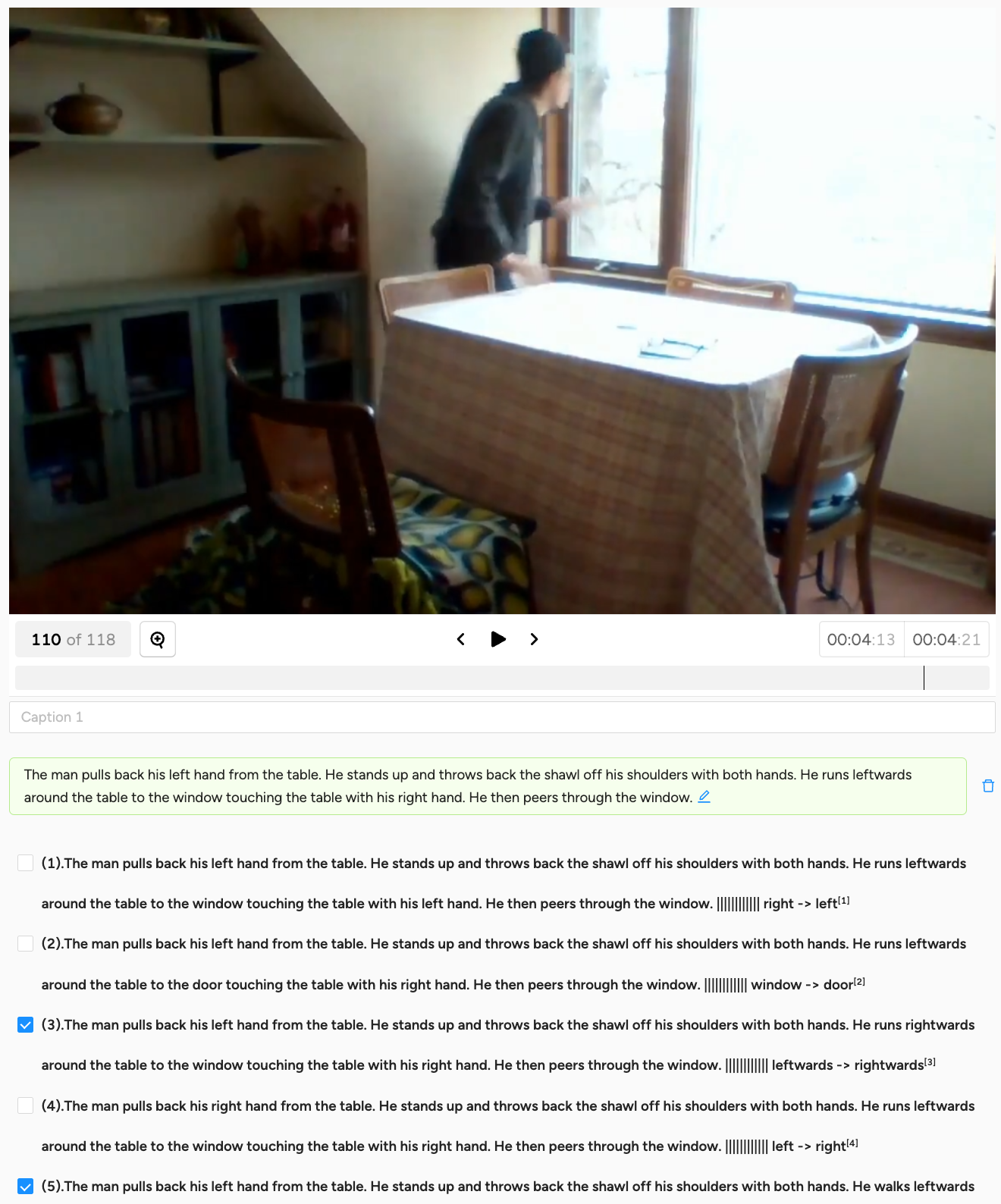}
    \caption{Negative caption annotation platform.}
    \vspace{-0.1em}
    \label{fig:example_neg_cap_choose}
\end{figure}

We first prompt LLMs (GPT-4o, Gemini, and Llama-3.1-405b) to get initial negative captions, and then ask authors to choose the negatives that can reflect the temporal dynamic. The visualization of the multi-choice platform in shown in Figure~\ref{fig:example_neg_cap_choose}.

\begin{table}[b]
    \centering
    \caption{\shortname{} performance of various multimodal generative models and embedding models under the binary QA accuracy (BA) and multiple binary QA settings (MBA) for short videos. The prefix ``T-" indicates \textbf{BA} performance for the annotated subset in our \shortname{}. We show the result with the best average MBA performance for each model with respect to the number of frames, denoted as \# Frames.
    } 
    \label{table:main_results_shortqa_BA}
    \resizebox{\textwidth}{!}{
    \begin{tabular}{lc|ccccccc|cc}
        \toprule

Model              & \# Frames         & T-ActivityNet & T-Charades & T-FineGym & T-Movie & T-Oops & T-COIN & T-EgoExo4D & BA & MBA  \\\midrule
Human Performance     & -    & \textbf{91.1 }       & \textbf{93.8}     &\textbf{ 77.0  }  & \textbf{93.1  }& \textbf{92.6} & \textbf{90.2} & \textbf{92.5 }  & \textbf{89.7} & \textbf{67.9}  \\
Random Chance     & -      &       50.0             &       50.0     &       50.0     &       50.0     &       50.0     &       50.0     &       50.0     &       50.0     &       50.0          \\ \midrule

\multicolumn{10}{c}{\textbf{Video Embedding Models: Text + Multiple Frames as Input}} \\ \midrule

XCLIP                 & 8    & 52.7        & 52.8     & 49.0    & 53.9  & 53.5 & 52.3 & 48.1   & 51.6 & 12.9  \\
ImageBind             & 2    & 52.9        & 52.6     & 47.5    & 55.4  & 56.8 & 52.4 & 53.4   & 53.0 & 14.0  \\
LanguageBind          & 8    & 56.5        & 50.1     & 48.2    & 55.8  & 55.1 & 51.1 & 52.8   & 52.8 & 14.5  \\

\midrule
\multicolumn{10}{c}{\textbf{Video Multimodal Generative Models : Text + Multiple Frames as Input}} \\ \midrule

GPT-4o                & 16   & 78.5        & 74.8     & 64.8    & 77.2  & 77.9 & 79.2 & 78.3   & 75.7 & 38.5  \\
Gemini-1.5-Pro        & 1FPS & 70.7        & 63.0     & 55.0    & 72.5  & 70.3 & 70.2 & 70.8   & 67.5 & 26.6  \\
Claude-3.5-Sonnet     & 8    & 68.5        & 62.4     & 62.7    & 68.2  & 64.2 & 65.4 & 66.8   & 65.5 & 23.6  \\
Qwen2-VL-72B & 32   & 76.6        & 74.5     & 65.4    & 79.8  & 77.7 & 77.2 & 79.7   & 75.8 & 38.3  \\
Qwen2-VL-7B  & 32   & 67.0        & 65.2     & 49.9    & 70.5  & 66.5 & 66.5 & 66.6   & 64.4 & 24.7  \\
LLaVA-OneVision-72B   & 8    & 76.0        & 70.4     & 59.3    & 76.1  & 75.2 & 73.5 & 74.9   & 72.1 & 33.0  \\
LLaVA-OneVision-7B    & 32   & 66.5        & 60.0     & 49.4    & 68.0  & 61.6 & 64.6 & 64.4   & 61.9 & 21.2  \\
LLaVA-NeXT-Video-34B  & 32   & 67.5        & 62.9     & 56.3    & 68.0  & 66.1 & 63.4 & 64.5   & 64.0 & 22.0  \\
LLaVA-NeXT-Video-7B   & 8    & 68.0        & 66.5     & 56.7    & 69.9  & 66.1 & 65.2 & 65.0   & 65.1 & 23.6  \\
InternLM-XC2.5        & 1FPS & 61.0        & 57.9     & 50.6    & 63.5  & 60.3 & 59.2 & 59.7   & 58.8 & 17.9  \\
VideoLLaVA            & 8    & 71.8        & 63.4     & 61.6    & 68.2  & 68.5 & 68.9 & 67.3   & 67.1 & 25.5  \\
MiniCPM-V2.6          & 1FPS & 66.1        & 59.6     & 54.1    & 68.0  & 63.1 & 62.7 & 62.7   & 62.3 & 21.4  \\
Phi-3.5-Vision        & 2    & 62.0        & 55.8     & 50.0    & 64.1  & 58.2 & 57.7 & 58.9   & 58.0 & 16.9  \\
MA-LMM                & 4    & 49.8        & 48.8     & 42.3    & 48.0  & 49.9 & 49.0 & 48.8   & 48.0 & 9.4   \\
\textit{$M^{3}$}        & 6    & 59.5        & 54.9     & 51.1    & 60.9  & 58.9 & 54.9 & 55.2   & 56.4 & 14.8  \\

 \midrule
\multicolumn{10}{c}{\textbf{Large Multimodal Models (LMMs): Text + 1 Frame as Input}} \\ \midrule
GPT-4o                & 1    & 69.1        & 67.1     & 64.8    & 71.0  & 71.9 & 71.0 & 74.0   & 70.0 & 28.4  \\
LLaVA-1.5-13B         & 1    & 57.6        & 54.3     & 51.9    & 56.8  & 53.2 & 58.1 & 57.8   & 55.7 & 13.1  \\
LLaVA-1.5-7B          & 1    & 64.2        & 58.6     & 55.7    & 61.0  & 57.5 & 62.7 & 63.9   & 60.5 & 18.3  \\
LLaVA-NeXT-34B        & 1    & 59.7        & 60.3     & 55.0    & 61.8  & 62.0 & 61.0 & 63.7   & 60.5 & 18.0  \\
Phi-3-Vision          & 1    & 57.4        & 54.5     & 45.2    & 57.5  & 52.8 & 55.8 & 58.9   & 54.4 & 15.1  \\

  \midrule
\multicolumn{10}{c}{\textbf{Large Language Models (LLMs): Text as Input}} \\ \midrule

GPT-4o                & 0    & 66.2        & 67.4     & 65.6    & 65.6  & 68.9 & 67.8 & 71.7   & 67.7 & 26.5  \\
Gemini-1.5-Pro        & 0    & 58.5        & 57.6     & 50.6    & 59.8  & 57.6 & 58.6 & 64.3   & 58.1 & 16.1  \\
Yi-34B                & 0    & 59.1        & 62.3     & 54.9    & 59.7  & 57.7 & 63.1 & 63.6   & 59.9 & 18.7  \\
Vicuna7b-1-5          & 0    & 49.7        & 49.5     & 50.2    & 50.7  & 50.5 & 50.0 & 52.1   & 50.5 & 10.4  \\
Flan-T5-XL            & 0    & 60.5        & 59.2     & 50.5    & 60.7  & 56.8 & 58.7 & 60.3   & 57.9 & 17.9  \\
Flan-T5-XXL           & 0    & 56.7        & 49.3     & 52.0    & 59.0  & 54.6 & 56.1 & 56.2   & 55.1 & 15.1 

\\ \bottomrule
\end{tabular}
}
\end{table}

\begin{table}[t!]
    \centering
    \caption{\shortname{} performance of various multimodal generative models and embedding models under \textbf{long video} understanding with binary QA accuracy (BA) and multiple binary QA accuracy (MBA). The \textbf{BA} performance under each dataset is also included. We show the result with the best average MBA performance for each model with respect to the number of frames, denoted as \# Frames. }
    \label{table:longvideo_BA}
    \resizebox{\textwidth}{!}{
    \begin{tabular}{lc|ccccc|cc}
        \toprule

Model                      & \# Frames      & T-ActivityNet & T-Charades & T-FineGym & T-COIN & T-EgoExo4D  & BA &  MBA \\\midrule
Random Performance   & -    & 50.0       & 50.0  & 50.0  & 50.0  & 50.0  & 50.0  & 50.0                      \\
\midrule
\multicolumn{9}{c}{\textbf{Video Embedding Models: Text + Multi-Frames as Input}} \\ \midrule

XCLIP                & 8    & 51.9        & 48.7     & 47.9    & 52.6 & 52.8   & 51.7            & 11.1                      \\
ImageBind            & 2    & 50.3        & 52.6     & 47.9    & 51.3 & 51.3   & 51.0            & 10.7                      \\
LanguageBind         & 8    & 51.9        & 46.4     & 48.2    & 52.0 & 53.7   & 51.6            & 12.0                      \\

 \midrule

\multicolumn{9}{c}{\textbf{Video Multimodal Generative Models : Text + Multi-Frames as Input}} \\ \midrule

GPT-4o               & 64   & 74.8        & 73.8     & 61.2    & 70.1 & 68.7   & 70.5            & 32.7                      \\
Gemini-1.5-Pro       & 1FPS & 67.0        & 61.6     & 60.6    & 65.9 & 65.9   & 65.2            & 24.7                      \\
Claude-3.5-Sonnet    & 8    & 66.8        & 63.7     & 56.7    & 63.1 & 66.6   & 64.6            & 24.5                      \\
Qwen2-VL-72B         & 8    & 68.5        & 59.6     & 62.5    & 59.6 & 70.0   & 64.7            & 26.2                      \\
Qwen2-VL-7B          & 32   & 60.7        & 58.0     & 49.9    & 59.8 & 61.9   & 59.7            & 18.8                      \\
LLaVA-OneVision-72B  & 4    & 67.0        & 63.5     & 61.2    & 55.8 & 69.3   & 63.4            & 23.8                      \\
LLaVA-OneVision-7B   & 32   & 60.0        & 53.6     & 57.6    & 53.2 & 59.8   & 56.9            & 16.2                      \\
LLaVA-NeXT-Video-34B & 4    & 59.4        & 63.0     & 57.6    & 59.5 & 61.4   & 60.3            & 20.0                      \\
LLaVA-NeXT-Video-7B  & 8    & 60.9        & 58.6     & 51.5    & 56.7 & 56.1   & 57.2            & 17.3                      \\
InternLM-XC2.5       & 1FPS & 59.6        & 58.9     & 57.0    & 54.9 & 52.8   & 55.8            & 15.6                      \\
VideoLLaVA           & 8    & 61.2        & 57.0     & 59.5    & 50.1 & 57.3   & 56.0            & 15.1                      \\
MiniCPM-V2.6         & 1FPS & 53.7        & 58.6     & 41.3    & 54.8 & 53.9   & 60.3            & 19.3                      \\
Phi-3.5-Vision       & 4    & 60.3        & 52.3     & 58.1    & 50.3 & 55.1   & 54.5            & 14.5                      \\
MA-LMM               & 4    & 47.4        & 51.7     & 36.4    & 50.1 & 51.2   & 47.1            & 9.2                       \\
\textit{$M^{3}$}       & 6    & 52.5        & 52.9     & 51.0    & 53.4 & 53.6   & 53.1            & 11.8                      \\


 \midrule
\multicolumn{9}{c}{\textbf{Large Multimodal Models (LMMs): Text + 1 frame as Input}} \\ \midrule

GPT-4o               & 1    & 67.6        & 64.3     & 62.8    & 65.9 & 62.0   & 64.7            & 24.5                      \\
LLaVA-1.5-13B        & 1    & 55.1	 & 52.3	 & 52.9	 & 55.0	 & 54.8	 & 54.5 & 	14.2                      \\
LLaVA-1.5-7B         & 1    & 51.2        & 53.4     & 51.5    & 51.8 & 56.2   & 53.2            & 12.3                      \\
LLaVA-NeXT-34B       & 1    & 60.6        & 60.8     & 57.0    & 59.8 & 61.8   & 60.5            & 19.9                      \\
Phi-3-Vision         & 1    & 56.9        & 53.9     & 52.1    & 55.6 & 57.6   & 56.0            & 15.6                      \\

\midrule
\multicolumn{9}{c}{\textbf{Large Larguage Models (LLMs): Text as Input}} \\ \midrule
GPT-4o               & 0    & 67.1        & 68.1     & 63.6    & 65.1 & 71.3   & 67.6            & 28.2                      \\
Gemini-1.5-Pro       & 0    & 62.8        & 59.4     & 55.6    & 60.7 & 65.7   & 62.2            & 21.2                      \\
Yi-34B               & 0    & 59.0        & 60.2     & 56.5    & 59.5 & 60.4   & 59.5            & 18.4                      \\
Vicuna7b-1-5         & 0    & 49.0        & 52.4     & 49.3    & 51.2 & 52.2   & 51.1            & 9.9                       \\
Flan-T5-XL           & 0    & 61.3        & 57.7     & 59.8    & 58.8 & 61.7   & 60.1            & 19.4                      \\
Flan-T5-XXL          & 0    & 59.4        & 53.6     & 59.5    & 56.3 & 56.5   & 56.9            & 16.7                     

\\ \bottomrule
\end{tabular}
}
\end{table}

\end{document}